\documentclass[review]{elsarticle}
\usepackage{hyperref}
\usepackage{float}
\usepackage{verbatim} 
\usepackage{apalike}
\usepackage{placeins}
\usepackage{multirow}
\usepackage{url}
\usepackage{csquotes}
\restylefloat{figure}
\usepackage{textcomp}
\usepackage{parskip}
\restylefloat{table}
\DeclareUnicodeCharacter{2212}{-}

\journal{Latex Template}

\biboptions{authoryear}

\begin{document}
\begin{frontmatter}
\begin{titlepage}
\begin{center}
\vspace*{1cm}

\textbf{ \large Profiling the news spreading barriers using news headlines}

\vspace{1.5cm}

Abdul Sittar$^{a}$ (abdul.sittar@ijs.si), 
Dunja Mladeni\'{c}$^{b}$ (dunja.mladenic@ijs.si), 
Marko Grobelnik$^c$ (marko.grobelnik@ijs.si) \\

\hspace{10pt}

\begin{flushleft}
\small  
$^a$ Jozef Stefan International Postgraduate School, Jamova Cesta 39, 1000 Ljubljana, Slovenia \\
$^b$ Jozef Stefan Institute, Jamova Cesta 39, 1000 Ljubljana, Slovenia \\
$^c$ Jozef Stefan Institute, Jamova Cesta 39, 1000 Ljubljana, Slovenia \\

\vspace{1cm}
\textbf{Corresponding Author:} \\
Abdul Sittar \\
Jozef Stefan International Postgraduate School, Jamova Cesta 39, 1000 Ljubljana, Slovenia \\
Tel: (+386) 41207252 \\
Email: abdul.sittar@ijs.si \\

\end{flushleft}        
\end{center}
\end{titlepage}

\title{Profiling the news spreading barriers using news headlines}

\author[label1]{Abdul Sittar  \corref{cor1}}
\ead{abdul.sittar@ijs.si}

\author[label2]{Dunja Mladeni\'{c}}
\ead{dunja.mladenic@ijs.si}

\author[label3]{Marko Grobelnik}
\ead{marko.grobelnik@ijs.si}

\cortext[cor1]{Corresponding author.}
\address[label1]{Jozef Stefan International Postgraduate School, Jamova Cesta 39, 1000 Ljubljana, Slovenia}
\address[label2]{Jozef Stefan Institute, Jamova Cesta 39, 1000 Ljubljana, Slovenia}
\address[label3]{Jozef Stefan Institute, Jamova Cesta 39, 1000 Ljubljana, Slovenia}

\begin{abstract}
News headlines can be a good data source for detecting the news spreading barriers in news media, which may be useful in many real-world applications. In this paper, we utilize semantic knowledge through the inference-based model COMET and sentiments of news headlines for barrier classification. We consider five barriers including cultural, economic, political, linguistic, and geographical, and different types of news headlines including health, sports, science, recreation, games, homes, society, shopping, computers, and business. To that end, we collect and label the news headlines automatically for the barriers using the metadata of news publishers. Then, we utilize the extracted commonsense inferences and sentiments as features to detect the news spreading barriers. We compare our approach to the classical text classification methods, deep learning, and transformer-based methods. The results show that the proposed approach using inferences-based semantic knowledge and sentiment offers better performance than the usual (the average F1-score of the ten categories improves from 0.41, 0.39, 0.59, and 0.59 to 0.47, 0.55, 0.70, and 0.76 for the cultural, economic, political, and geographical respectively) for classifying the news-spreading barriers.
\end{abstract}

\begin{keyword}
News spreading barriers\sep Profiling news spreading barriers\sep Commonsense inferences \sep sentiment analysis \sep Economic barrier \sep Political barrier \sep Cultural barrier\sep Linguistic barrier\sep Geographical barrier\sep
\end{keyword}

\end{frontmatter}

\section{Introduction}\label{sec:Introduction}

\begin{figure*}[!htb]
\centering
        \includegraphics[width=0.48\textwidth]{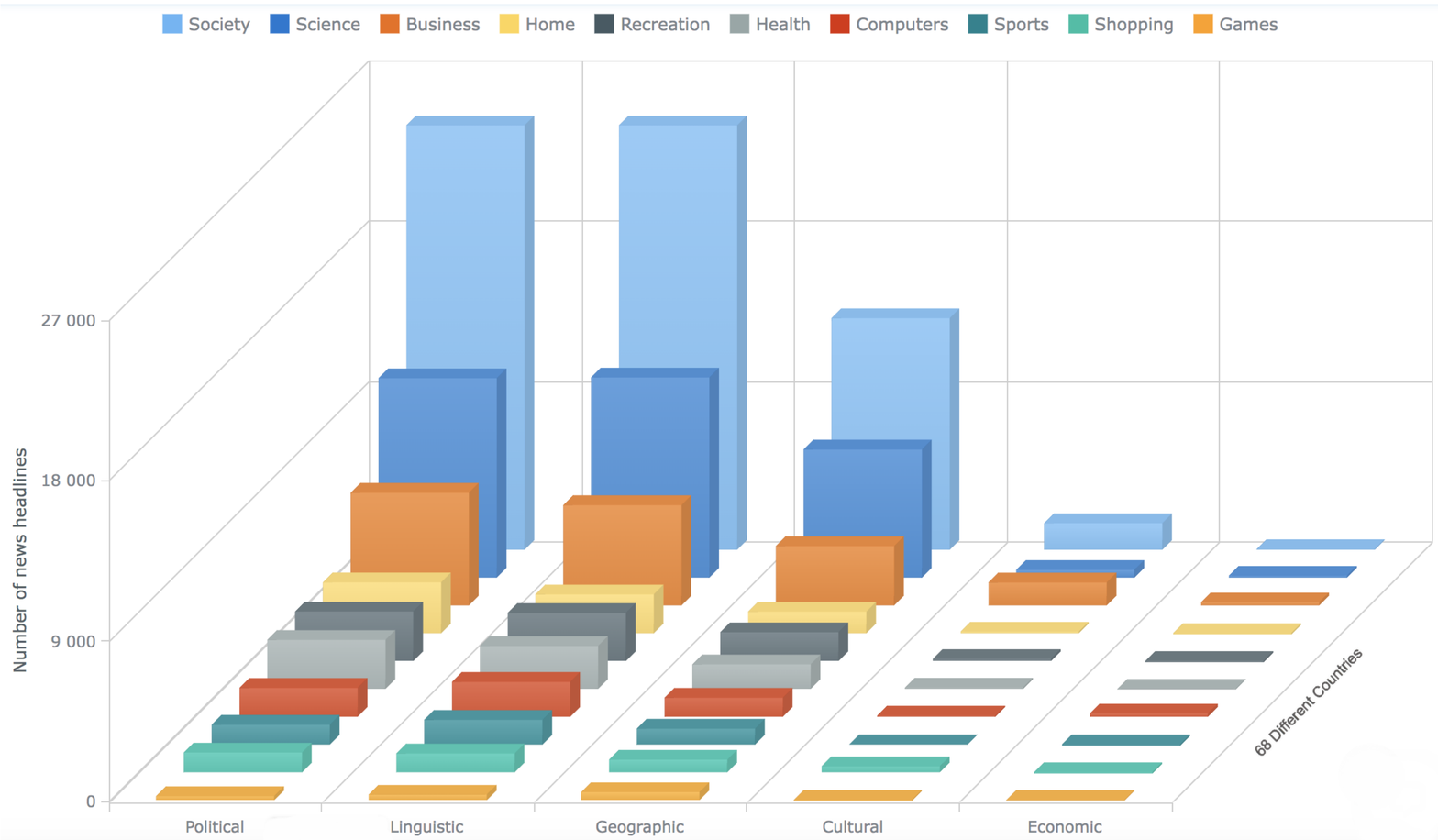}
        \includegraphics[width=0.45\textwidth]{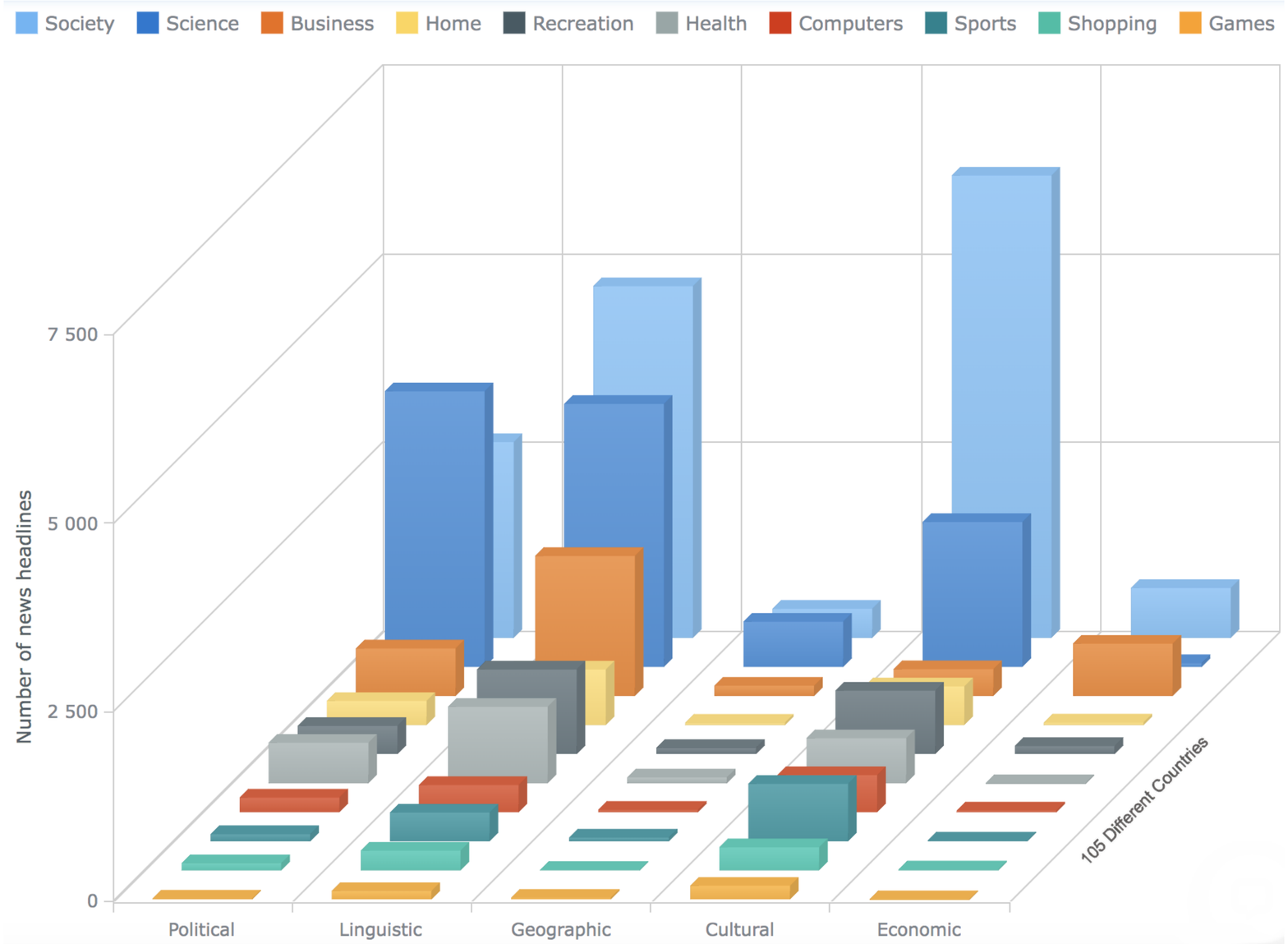}
        \includegraphics[width=0.45\textwidth]{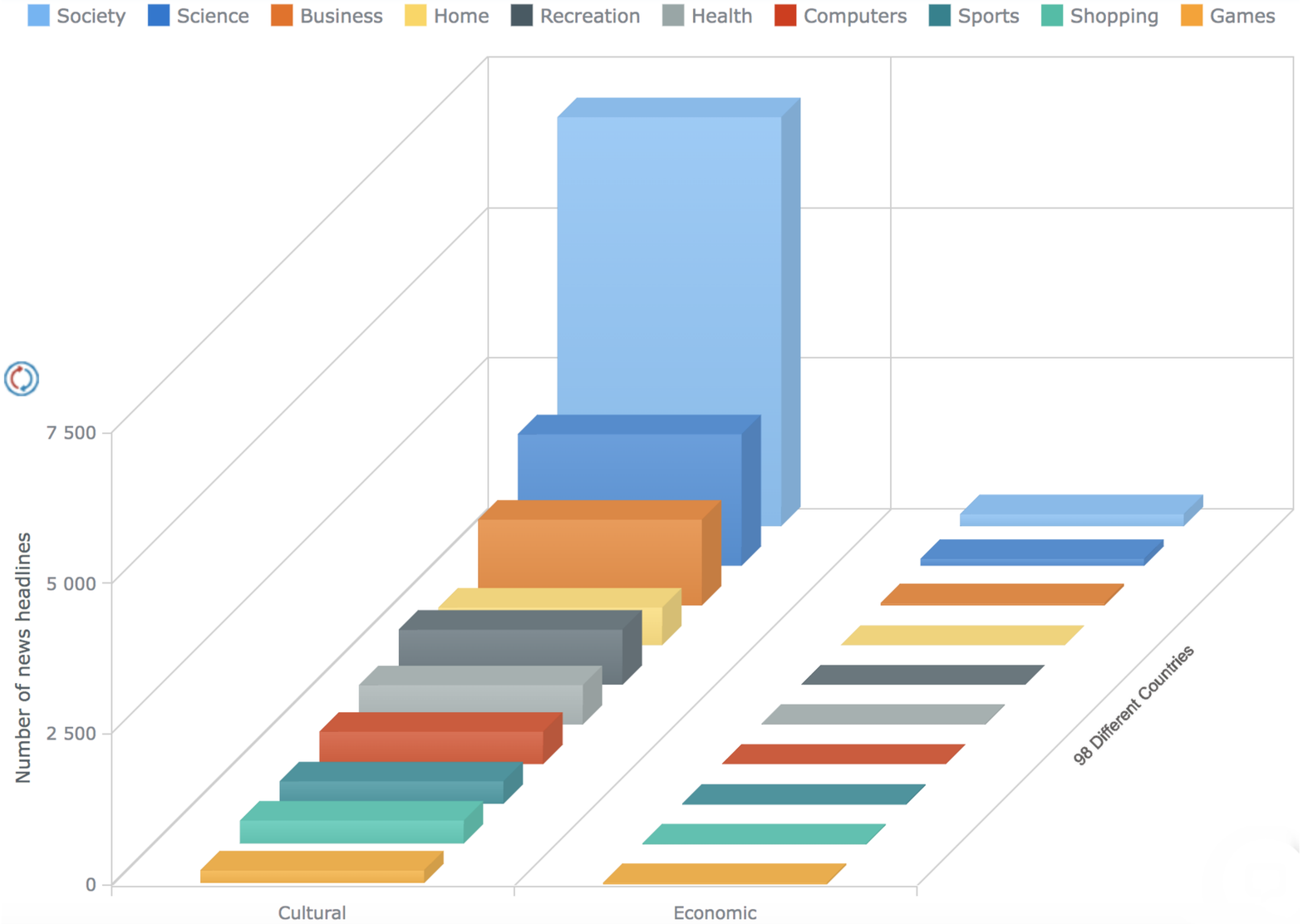}
        \caption{\bf The bar charts show the statistics about the news articles that have the labels "Information-crossing", "information-not-crossing", and "unsure" respectively (from left to right) for all the ten different categories. The x-axis shows the different barriers, the y-axis shows the count of the news articles, whereas the bars on the z-axis represent ten different categories (see Section \ref{sec:datadescription}).}
        \label{fig:labeleddata}
\end{figure*}

Event-centric news spreading comes across many barriers because the storylines of the news are anchored to time, place and entities \cite{ref:sittar2022analysis, ref:rospocher2016building}. In this paper, the term barriers refer to the bridges that are in place between different societies, nations, and countries while transferring information to each other. The classification of such barriers can be useful in the context of numerous real-world applications, such as trend detection, and content recommendations for readers and subscribers \cite{ref:gulla2017adressa, ref:heydari2015detection}. Thus, it is highly important to classify the barriers to massive news spreading related to different events.
\newline
\newline
Culture is multifaceted, subsuming behaviors, values, and attitudes that are dominant and unique to a particular group of people. News media has a strong relationship with many macro-level factors in society, ranging from the economy, governments, the public, and other organizational structures \cite{ref:ng2021diversity}. In a cultural barrier, media diversity provides different opinions and perspectives across different cultures \cite{ref:d2009belgium}. The publishing language of a news media also influences the diffusion of news about global and local events \cite{ref:wright2022social} where we can say that there is a language barrier. Likewise, the political alignment of news publishers has a direct relation with the published content and it is called by \cite{ref:sittar2022analysis} as a political barrier. Another contextual variable that has a direct relation with different types of news is the economic situation (we call it the economic barrier) surrounding the news publisher. Since the economic differences in the living style affect the need, the news is likely to propagate according to the need of locals. 
\newline
\newline
Another important variable in this context is news sentiments about different events across different locations. Several studies use a sentiment from textual data including social media, and news articles to forecast financial variables \cite{ref:barbaglia2022forecasting, ref:consoli2022fine, ref:kumbure2022machine}. The sentiment of the news plays an important role in news spreading as \cite{ref:bustos2011pricing} found that the price movement in Stock Exchange has a direct relation with the news spreading patterns. Similarly, news about global events has different sentiment polarity across the geographical barrier. \cite{ref:moreo2012lexicon} calls it the popularity measurement of news in a global context. Market behavior is also predictable through sentiments \cite{ref:godbole2007large, ref:shah2018predicting} as well as sentiment can vary by demographic group, news sources and geographic locations \cite{ref:mehler2006spatial}. 
\newline
\newline
When it comes to news headline, it reflect the vital information of news articles. It reduces the interpretation time and effort of reading the whole article \cite{ref:shrawankar2016construction}. The first thing in the news article is its headline which makes the first and foremost impression on the news readers. Plenty of news articles are published every day and spread via news and social media\cite{ref:gravanis2019behind, ref:gabielkov2016social, ref:nassirtoussi2015text}. These headlines have different emotional scores with a negative, positive or neutral polarity which directly impacts the readers' actions \cite{ref:aslam2020sentiments}. 
\newline
\newline
Barrier classification with news headlines is a challenging task due to incorporating insufficient information as well as misinformation in the headlines. News coverage of different fields including sports, health, and computers has different impact levels. We focus on five different types of barriers including cultural, political, linguistic, economic, and geographic as these are important barriers that can influence news spreading \cite{ref:sittar2022analysis}. In this study, we assume that common sense-based semantic knowledge and sentiments of news headlines will help to classifying the news-spreading barriers. We are interested in exploring the variations in sentiments across different barriers where news headlines belong to different events. We explore a range of different commonsense descriptions generated by the Natural Language Processing Knowledge Inference Tool 
\cite{ref:Ismayilzada2022kogito}. In addition, we present an approach to barrier classification that aims to classify barriers across the news. This approach combines information based on news headlines, their inferences, and its sentiment.
\newline
\newline
More specifically, we focus on the following research questions:
\begin{description}
  \item  RQ1: Do the sentiments of the news headlines of different topics vary across the different barriers?
  \item  RQ2: What are the properties (statistics and ratio) of the common sense knowledge relations in news headlines to different topics?
  \item  RQ3: Which classification methods (classical or deep learning methods or transformers based methods) yield the best performance to barrier classification task?
\end{description}

\noindent The contributions of this research can be summarized as follows:

\begin{enumerate}
  \item A novel approach to information barrier annotation based on news meta-data.
   \item A dataset for the barrier classification in the news that has been labeled automatically using the metadata and the semantic similarity. 
  \item An approach to classification of news spreading barriers based on semantic knowledge including a wide range of common sense inferences and sentiments of news headlines.
\end{enumerate}

The rest of the paper is organized as follows: Section \ref{sec:RelatedWork} reviews the related work on barrier classification; Section \ref{sec:approach} presents the approach; Section \ref{sec:datadescription} describes the benchmark dataset construction; Section \ref{sec:MethodsforAutomaticBarrierProfiling} discusses the experimental results; 
Section \ref{sec:conclusions} concludes the paper and highlights the theoretical and practical implications of our study.
\section{Related Word}\label{sec:RelatedWork}
In this section we present the related work on news spreading barriers and the role of semantic knowledge and sentiments of the news headlines for different tasks.
\subsection{News spreading barriers}
Effective dissemination is the key to bridging the gap in information spreading. For the scientists and the practitioners, it is necessary to participate in explicit, accurate, and unbiased dissemination of their respective areas of expertise to the public \cite{ref:kelly2019spreading}. The result of communication is not only situation-specific but also inherently culturally bound because it is entrenched in human acts with intentions, interests, and wants as well as larger institutional, social, and cultural systems \cite{ref:jiang2020relying}. Culture-specific ideology is defined as the values, beliefs, attitudes, or interests expressed in a source text that is associated with a particular culture or source and that may be viewed as undesirable or incompatible with the dominant values, beliefs, attitudes, or interests of another culture or subculture. It defines the strategies adopted by text producers in bridging the divides in global news transmission. According to MCNelly's theory, the more distance an intermediary communicator has to travel before learning about a news occurrence, the less personally invested he is in it and the more he considers its "marketability" to editors or readers \cite{ref:vuorinen1994crossing}. It has been said that countries with close distances share culture and the news reporting on the same events will not differ due to ideology, culture, and geopolitics \cite{ref:segev2015visible, ref:ma2017does}. Countries that share a common culture are expected to have heavier news flow between them when reporting on similar events \cite{ref:wu2007brave}. There are many quantitative studies that found demographic, psychological, socio-cultural, source, system, and content-related aspects \cite{ref:al2017impact}. The role of content is an essential research topic in news spreading. Media economics scholars especially showed their interest in a variety of content forms since content analysis plays a vital role in individual consumer decisions and political and economic interactions \cite{ref:fico2008content}. In content, a frame is a means to highlight certain elements of a seen reality in a communication text so as to support a specific problem definition, causal interpretation, moral assessment, and/or therapy proposal for the thing being described. There are four places where frames can be found during communication: the text, the recipient, the communicator, and the culture \cite{ref:reese2007framing}. The inverted pyramid reporting method, where the most significant facts are presented in order of importance, is a key component of news framing. Bias in the news can manifest in a variety of ways, these include "source bias," "unbalanced presentation of contested themes," and "frequent usage of packaged formula" \cite{ref:walter2019news}. Scheufele identifies five factors that influence how journalists frame news. These include societal expectations and ideals, organizational demands and restrictions, pressure from interest groups, journalistic practices, and journalists' ideological or political leanings \cite{ref:obijiofor2010press}. A vast body of literature exists on how the news media frame the news events and consequently influence public perception of those events \cite{ref:lamidi2016newspaper}. Existing literature posit that framing is often used intentionally for the purpose of changing the perception of content and to cater to this, different computational methods have been applied \cite{ref:king2017news, ref:sheshadri2021detecting}.
\subsection{Inference-based semantic knowledge}
COMmonsense Transformers (COMET) is an automatic construction of common-sense knowledge bases. It is a framework for adapting the weights of language models to learn to produce novel and diverse common-sense knowledge tuples \cite{ref:bosselut2019comet}. Abductive natural language inference can be used to interpret between the lines in natural language \cite{ref:bhagavatula2019abductive}. Inferences allow us to connect pieces of knowledge to reach the new conclusion. Human performs natural language inference based on a vast amount of external knowledge about language and the world. In order to comprehend human language, machines first need linguistic knowledge i.e., knowledge about the language. This includes the understanding of word meanings, grammar, syntax, semantic, and discourse structure. Having linguistic knowledge gives a human or machine the basic capabilities of understanding language, and is a required property of virtually any NLP system, even those not created for NLI tasks. Common knowledge refers to well-known facts about the world that are often explicitly states \cite{ref:cambria2011isanette}. This kind of knowledge is often referred to in human communication \cite{ref:cambria2014senticnet}. Some types of common knowledge may be domain-specific. While domain-specific knowledge is obviously useful for domain-specific applications, much of this knowledge may not be needed for general-purpose communication with humans. Common-sense knowledge, on the other hand, is typically unstated as it is considered obvious to most humans and consists of universally accepted beliefs about the world. Commonsense knowledge provides a deeper understanding of language. While it is rarely referred in language, humans rely on it in communication \cite{ref:gao2016physical}, as it is required to reach a common ground. It consists of everyday assumptions about the world, and is generally learned through one's own experience with the world, but can also be inferred by generalizing over common knowledge. While common knowledge can vary by region, culture and other factors, we expect that commonsense knowledge should be roughly typical to all humans \cite{ref:davis2017first}. To tackle the challenging benchmark tasks, many computational models have been developed. These range from earlier symbolic and statistical approaches to recent approaches based on deep neural networks. Explicit textual content is used for different task such as hate speech detection systems and the primary challenges for statistical and neural classifiers is the infer the implicit messages in text. Recent works have highlighted the needs to use implicit messages to detect textual content \cite{ref:elsherief2021latent}. Similarly external knowledge base was used with transformer to perform emotion recognition, bias prediction \cite{ref:ghosal2020cosmic, ref:swati4114271ic}. Semantic knowledge also proved to enhance the existing model to learn a general representation \cite{ref:razniewski2021information}. There are many example of recommendation systems that utilize the semantic knowledge consisting on several attributes and multi-model knowledge \cite{ref:lei2021suggested, ref:zhou2020recommending}. Taking the semantic information through knowledge graphs is also one of the best way to associate semantic information to the data for different tasks \cite{ref:colon2021combining}. Common sense knowledge consists on many spatio-temporal features including spatial, physical, temporal, and psychological aspects of everyday life. It has proved to be crucial for many NLP tasks including dialogue understanding and generation, event prediction, and question answering \cite{ref:fang2021discos}. For the development of new approaches to address different tasks, one of the critical task is creating benchmark datasets to evaluate the approaches \cite{ref:storks2019recent}.

\subsection{Sentiments as semantic knowledge}
Sentiment classification of news deals with the identification of positive and negative news that can be used to predict the trends related to different tasks \cite{ref:yazdani2017sentiment}. Sentiment of the news has already been used for news classification as well as other features including entities, special phrases \cite{ref:hui2017effects, ref:demirsoz2017classification}. In the task of sentiment classification approaches, DistilBERT can transfer basic semantic understanding to further domains and lead to greater accuracy than the baseline TFIDF \cite{ref:dogra2021analyzing}. For the task of fake news detection, textual content of news along with the headline have been used to extract the features \cite{ref:cui2019same}. \cite{ref:taj2019sentiment} used dictionary-based and corpus-based methods for sentiment analysis of news related to business, entertainment, politics, sport, and technology. \cite{ref:li2017sentiment} have used sentiments along with bag-of-features to predict the stock market prediction. prove that sentiments of fake news increase the accuracy of fake news detection \cite{ref:bhutani2019fake} and there exists a strong relationship between news and its sentiments such as negative emotions tend to spread fast \cite{ref:ajao2019sentiment}.

\section{Approach Description}\label{sec:approach}
\begin{figure*}
\centering
\includegraphics[keepaspectratio=true,scale=0.55]{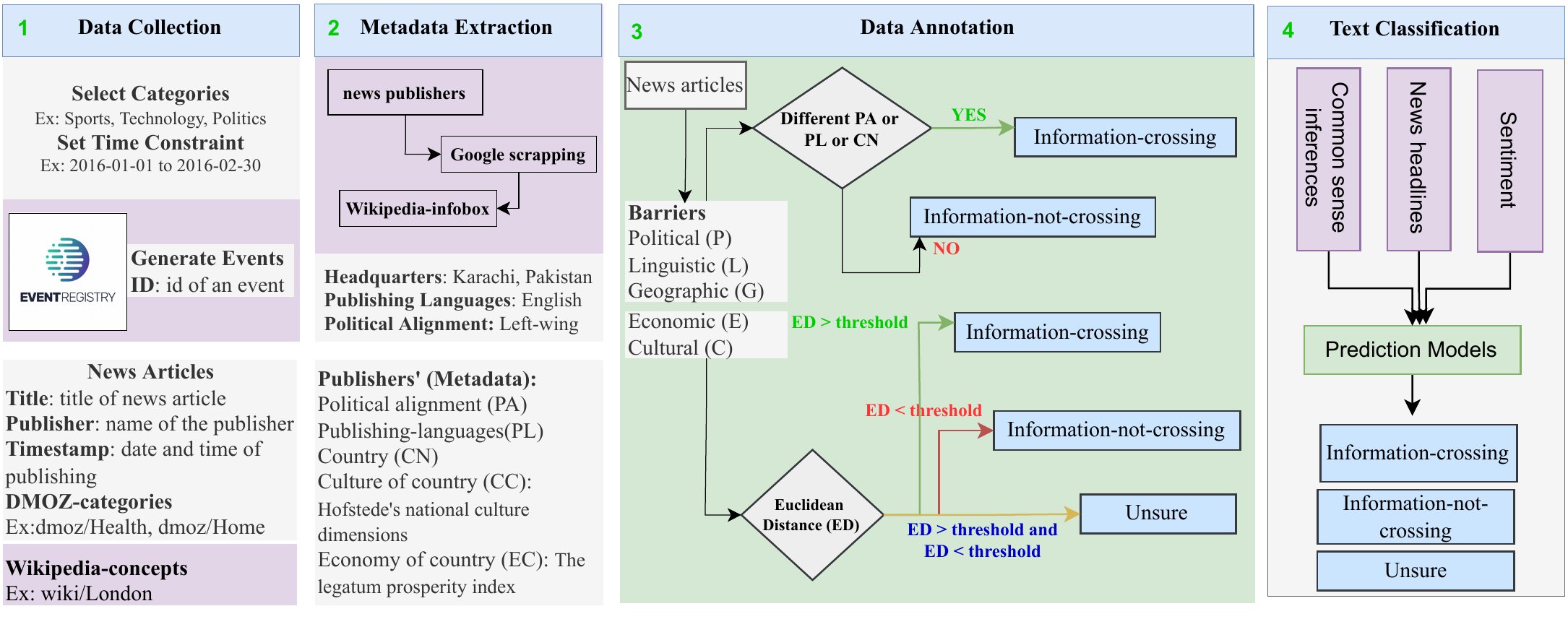}
\caption{\bf An approach to automatic barrier profiling based on the news meta-data. Data extraction from the Event Registry is the first step. Meta-data extraction through Google and Wikipedia scrapping is the second step. The third step is to label the news articles after calculating the euclidean distances. The classification with the classical machine learning models, deep learning, and transformer-based methods is performed in the last step.}
\label{fig:approach}
\end{figure*}

In order to perform the classification of news published across barriers (geographical, cultural, economic, etc.) and by that attempt to recommend and identify trends of news spreading belonging to different categories, some methodological considerations are necessary. 
\newline
\newline
This research article presents a novel approach to barrier annotation utilizing news meta-data and approach to news classification utilizing inferences based semantic knowledge, as shown in Figure \ref{fig:approach}. In the first step, we execute a query that extracts the news articles from the Event Registry belonging to different categories (business, computers, games, health, home, recreation, science, shopping, society, and sports) and published within a certain time span - in our case between 2016-2021 (see Subsection \ref{sec:datadescription}). Then we parse and save these news articles along with the source information such as publishers' names and publishing dates. 
\newline
\newline

In the second step, we extract meta-data related to news publishers via searching the news publishers' on Google and extracting their Wikipedia links. Using these links, we obtain the necessary information from Wikipedia-infobox \cite{ref:sittar2022analysis}. We use Bright Data service \footnote{\url{https://brightdata.com/}} to crawl and parse Wikipedia-Infobox.
\newline
\newline
In the third step, we perform the annotation of news articles. To label the news articles, we set the annotation guidelines (see Subsection \ref{sec:datadescription}). For cultural and economic barriers, we assign ternary labels to news articles whereas, for linguistic, geographical, and political barriers, we assign binary labels to news articles. 
\newline
\newline
In the fourth step, we conduct a detailed analysis of the sentiments of the news headlines for each category (see Figure \ref{fig:sentimentComparision}) and provide a list of comprehensive trends of sentiments across different categories and barriers (see Figure \ref{fig:sentimentalTrends}). Next, we extract semantic knowledge through the inference-based model COMET \cite{ref:bosselut2019comet}(see Figure \ref{fig:networkDiag}). We analyze the properties of the inferences relations to the news headlines of different topics (see Figure \ref{fig:venn_result-inferences}). Afterward, we conduct experiments comparing machine learning state-of-the-art (LR, NB, SVC, kNN, and DT), deep learning (LSTM) and transformer-based method (BERT) using a combination of news headlines with inferences based semantic knowledge and its sentiments. The results are presented in Section \ref{subsec:comAnaTenCates}, \ref{subsec:comAnaThreAlgos} showing the performance of different features and different methods. The source code of this approach is available in the GitHub repository \footnote{\url{https://github.com/abdulsittar/BC-Inferences-Sentiments.git}}. 
\section{Benchmark dataset construction}\label{sec:datadescription}
We collected the news articles reporting on different events published between 2016-2021 in the English language using Event Registry \cite{ref:leban2014event} APIs \footnote{\url{https://github.com/EventRegistry/event-registry-python/blob/master/eventregistry/examples/QueryArticlesExamples.py}}. The dataset consists of approximately 1.7 million news articles. Each news article belongs to a different category (see Table \ref{tab:totalfields}). Each news article consists of a few attributes: title, body text, name of the news publisher, date and time of publishing, event-ID, DMOZ-categories, and Wikipedia concepts. 
\begin{table}
\centering
\caption{\bf The table shows the statistics before and after filtering the news articles based on common sense knowledge extraction and data annotation for the ten categories (business, computers, games, health, home, recreation, science, shopping, society, and sports) of five barriers.}
\scalebox{0.90}{
\label{tab:totalfields}
\begin{tabular}{lccccc}
\textbf{Categories} & \multicolumn{1}{l}{\textbf{Cultural}} & \multicolumn{1}{l}{\textbf{Economic}} & \multicolumn{1}{l}{\textbf{Geographic}} & \multicolumn{1}{l}{\textbf{Linguistic}} & \multicolumn{1}{l}{\textbf{Political}}  \\ 
\hline
\textbf{Business}   & 37827                                 & 23641                                 & 40084                                   & 51334                                   & 28882                                   \\
\textbf{Computers}  & 12148                                 & 1475                                  & 10385                                   & 39777                                   & 9105                                    \\
\textbf{Games}      & 3128                                  & 512                                   & 2873                                    & 2757                                    & 1754                                    \\
\textbf{Health}     & 14735                                 & 1880                                  & 14981                                   & 19770                                   & 14743                                   \\
\textbf{Home}       & 13890                                 & 1989                                  & 13467                                   & 20109                                   & 13536                                   \\
\textbf{Recreation} & 2105                                  & 3087                                  & 17026                                   & 21492                                   & 16014                                   \\
\textbf{Science}    & 80256                                 & 10031                                 & 83594                                   & 83538                                   & 82563                                   \\
\textbf{Shopping}   & 6426                                  & 863                                   & 6543                                    & 8783                                    & 5398                                    \\
\textbf{Society}    & 159170                                & 20461                                 & 190606                                  & 218048                                  & 144744                                  \\
\textbf{Sports}     & 23641                                 & 1076                                  & 8292                                    & 10279                                   & 5667                                    \\
\multicolumn{6}{c}{\textbf{After common sense extraction and data annotation}}                                                                                                                                                    \\ 
\hline
\textbf{Business}   & 3455                                  & 1015                                  & 3550                                    & 7974                                    & 7037                                    \\
\textbf{Computers}  & 913                                   & 310                                   & 1181                                    & 2194                                    & 1895                                    \\
\textbf{Games}      & 186                                   & 68                                    & 549                                     & 504                                     & 316                                     \\
\textbf{Health}     & 1159                                  & 90                                    & 1533                                    & 3295                                    & 3368                                    \\
\textbf{Home}       & 1065                                  & 86                                    & 1321                                    & 2796                                    & 3258                                    \\
\textbf{Recreation} & 1695                                  & 161                                   & 1783                                    & 3697                                    & 3236                                    \\
\textbf{Science}    & 4377                                  & 378                                   & 7877                                    & 14665                                   & 14925                                   \\
\textbf{Shopping}   & 513                                   & 42                                    & 796                                     & 1685                                    & 1287                                    \\
\textbf{Society}    & 14238                                 & 1003                                  & 13472                                   & 28431                                   & 28447                                   \\
\textbf{Sports}     & 1533                                  & 39                                    & 1021                                    & 2054                                    & 1289                                    \\
\hline
\end{tabular}}
\end{table}
\FloatBarrier

\begin{figure}
\centering
\includegraphics[keepaspectratio=true,scale=0.35]{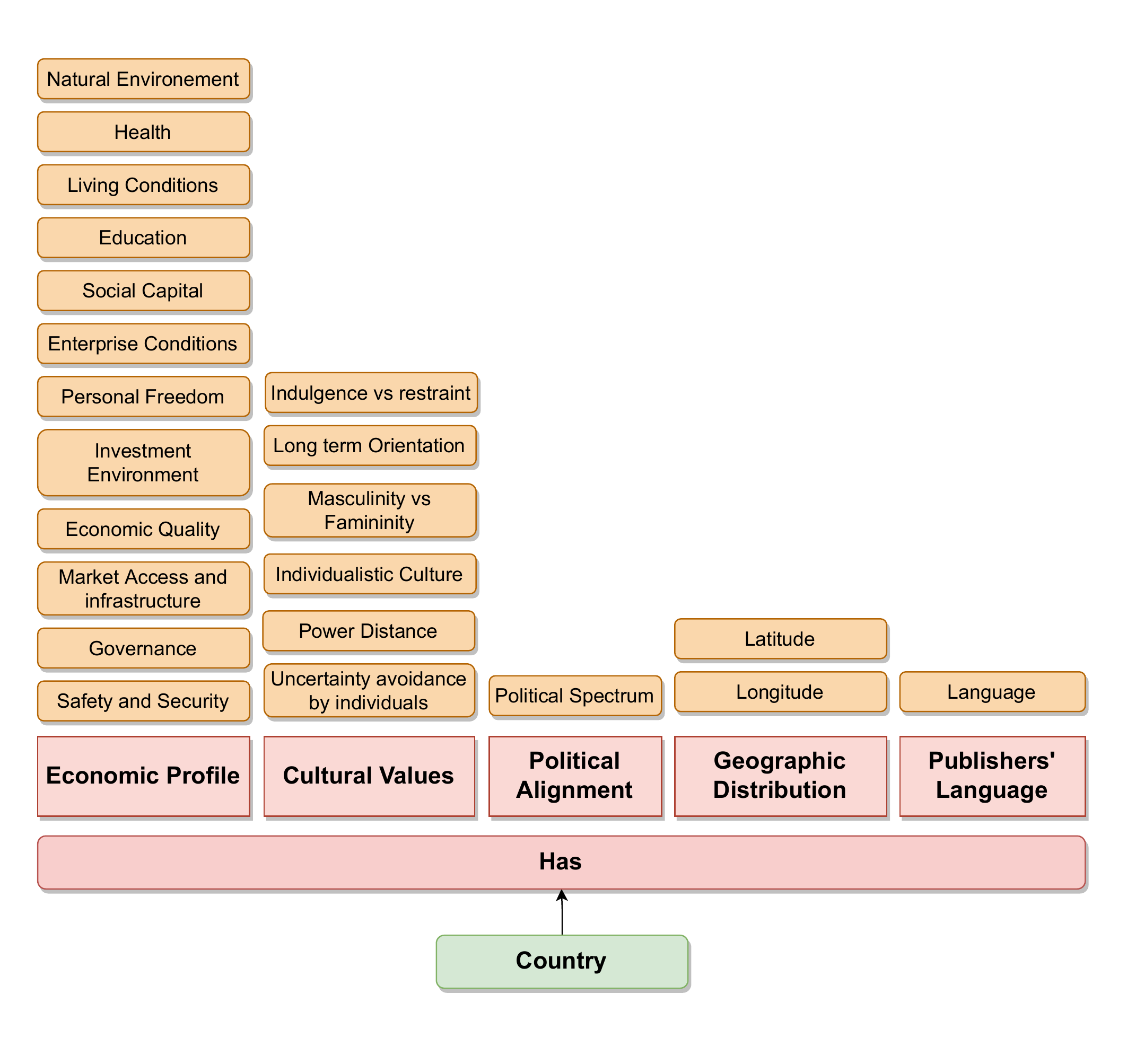}
\caption{Metadata for the five barriers (cultural, economic, geographical, linguistic, and political).}
\label{fig:schema}
\end{figure}

A few attributes are self-explanatory such as title, body text, name of the news publisher, and date and time of publishing. An event-id represents a unique number that is associated with all the news articles that belong to the same event. The DMOZ-categories represent topics of the content/news article. It is a project that has a hierarchical collection of web page links organized by subject matters \footnote{\href{https://dmoz-odp.org/}{https://dmoz-odp.org/}}. Around 50,000 categories are used by the Event Registry (top 3 layers of the DMoz taxonomy)  \footnote{\href{https://eventregistry.org/documentation?tab=terminology}{https://eventregistry.org/documentation?tab=terminology}}. The statistics of all the categories for all the five barriers are presented in the Table \ref{tab:totalfields}. The Wikipedia concepts are used as a semantic annotation for news articles and can represent entities (locations, people, organizations) or non-entities (things such as personal computers, and toys). In Event Registry, Wikipedia's URLs are used as concept URIs. 
\newline
To fetch the metadata for each barrier, the essential thing is the news publisher's headquarters name. For each news publisher we get this information from Wikipedia-infobox. We used the Bright Data service \footnote{\href{https://brightdata.com/}{https://brightdata.com/}} to crawl and parse Wikipedia-Infobox for almost more than 10,000 news websites. We retrieved the country name of the news publishers headquarters name. For the economical barrier, we fetched the economical profile for each country using \enquote{The Legatum Prosperity Index} \footnote{\href{https://www.prosperity.com/}{https://www.prosperity.com/}} as done by \cite{ref:sittar2022analysis}. It has twelve dimensions that represent different economical aspects (see Figure \ref{fig:schema}). For the cultural barrier, we calculated differences among different regions using six Hofstede’s national culture dimensions (HNCD) (see Figure \ref{fig:schema}). For the economic and cultural barrier, we calculated the euclidean distance among all the countries (for the economic barrier using the economical profile, and for the cultural barrier using the HNCD). Two countries have been labeled as: "information-not-crossing" if the distance score was $\leq$ 0.1, "unsure" if the distance score was $>$ 0.1 and $\leq$ 0.4 , "information-crossing" if the distance score was $>$ 0.4. For the geographical barrier, we stored general latitude and longitude. For the political barrier, we utilize the political ideology/alignment of the newspaper/magazine that we determined based on Wikipedia-infobox at their Wikipedia page \cite{ref:sittar2022political}. The statistics about the labeled dataset are presented in Figure \ref{fig:labeleddata} and Table \ref{tab:totalfields}. Data can be found in the GitHub repository \footnote{\url{https://github.com/abdulsittar/BC-Inferences-Sentiments.git}}.

\FloatBarrier

We set the following annotation questions based on the definitions above in order to classify the barriers to news spreading.
\begin{itemize}
\setlength\itemsep{0em}
\item Q1: \emph{Do all the news articles reporting on an event, publish from a particular/same geographical location?}
\item Q2: \emph{Do all the news articles reporting on an event, publish from the locations having equal economic prosperity?}
\item Q3: \emph{Do all the news articles reporting on an event, publish from a particular/same locations having equal cultures?}
\item Q4: \emph{Do all the news articles reporting on an event, publish from the sources with a particular/similar political class?}
\item Q5: \emph{Do all the news articles reporting on an event, publish by the newspapers where the publishing language were same?}
\end{itemize}

Question 1~(Q1) intends to identify whether the news was published across different geographical places or not. The question is answered "Yes" for all the news articles reported on an event if they are published from one country otherwise "No". Question 2~(Q2) intends to identify whether the news was published across different economies or not. The economic similarity has been calculated using euclidean distance. The question is answered with "information-crossing" for all the news articles reported on an event if they are published from countries with similar economic situations. The question is answered with "unsure" for all the news articles reported on an event if at least one of the news articles published from a country that is labeled with "unsure" (see Figure \ref{fig:schema}) otherwise "information-not-crossing". Question 3~(Q3) intends to identify whether the news was published across different cultures or not. The question is answered with "information-crossing" for all the news articles reported on an event if they are published from countries with a similar culture. The question is answered with "unsure" for all the news articles reported on an event if at least one of the news articles is published from a country that is labeled with "unsure" otherwise "information-not-crossing". The cultural similarity has been calculated using the euclidean distance (see Section \ref{sec:datadescription}). Question 4~(Q4) intends to identify whether the news was published in newspapers with the same political alignments or not. The question is answered "Yes" for all the news articles reporting on an event if they are published in the newspapers following similar political alignments otherwise "No". Question 5~(Q5) intends to identify whether the news was published in the newspapers where the publishing language was the same or not. The question is answered "Yes" for all the news articles reporting on an event if they are published from different newspapers where the publishing language was the same otherwise "No".

\subsection{Annotated dataset} \label{subsec:annodata}
Initially, we collected almost approximately 1.7 million news articles. After filtering the news based on the unavailability of the metadata information the news articles were limited to a few thousand news articles. Similarly based on not having no common sense inferences, the news article was reduced to a few thousand articles. The statistics of the news belonging to the ten categories across the five barriers are presented in Table \ref{tab:totalfields}. The dataset is available in the GitHub repository \footnote{\url{https://github.com/abdulsittar/BC-Inferences-Sentiments.git}}. Labels for the five types of barrier annotations are derived:
\begin{itemize}
\setlength\itemsep{0em}
\item economic barrier classes: \textit{information-not-crossing}, \textit{unsure}, and \textit{information-crossing}.
\item cultural barrier classes: \textit{information-not-crossing}, \textit{unsure}, and \textit{information-crossing}.
\item geographical barrier classes: \textit{Not-crossed-GB}, and \textit{Crossed-GB}.
\item Political barrier classes: \textit{Not-crossed-PB}, and \textit{Crossed-PB}.
\item Linguistic barrier classes: \textit{Not-crossed-LB}, and \textit{Crossed-LB}.
\end{itemize}

\begin{table*}
\centering
\caption{\bf This table shows the examples of annotation for all the five types of barriers. The annotation is performed using the meta-data shown in Figure \ref{fig:schema}.}
\scalebox{0.40}{
\label{tab:annotationExamples}
\begin{tabular}{|l|l|l|l|l|l|} 
\hline
\begin{tabular}[c]{@{}l@{}}\textbf{Barrier~}\\\textbf{(Category)}\end{tabular}    & \textbf{Time}                                                                                            & \textbf{Title}                                                                                                                                                                                                                                             & \begin{tabular}[c]{@{}l@{}}\textbf{Location/Publisher/}\\\textbf{Language}\end{tabular}                                                  & \textbf{Meta-Data}                                                                        & \textbf{Class}            \\ 
\hline
\begin{tabular}[c]{@{}l@{}}\textbf{cultural }\\\textbf{(Games)}\end{tabular}      & \begin{tabular}[c]{@{}l@{}}2016-07-18T19:48:00Z\\2016-07-18T22:04:00Z\end{tabular}                       & \begin{tabular}[c]{@{}l@{}}Trump aims to show his softer side at Cleveland convention\\Uproar at Republican convention as anti-Trump delegates \\revolt\end{tabular}                                                                                       & \begin{tabular}[c]{@{}l@{}}Ireland~(irishtimes.com)\\Thailand~(bangkokpost.com)\end{tabular}                                             & Same Culture                                                                              & information-not-crossing  \\ 
\cline{2-6}
                                                                                  & \begin{tabular}[c]{@{}l@{}}2016-04-16T22:23:00Z\\2016-04-18T16:02:00Z\end{tabular}                       & \begin{tabular}[c]{@{}l@{}}Another small victory for Cruz\\Romney: 3-man race throws Trump the nomination\end{tabular}                                                                                                                                     & \begin{tabular}[c]{@{}l@{}}New Zealand (odt.co.nz)\\United States (wcyb.com)\end{tabular}                                                & Different Culture                                                                         & un-sure                   \\ 
\cline{2-6}
                                                                                  & \begin{tabular}[c]{@{}l@{}}2016-05-25T05:19:00Z\\2016-05-25T11:40:00Z\end{tabular}                       & \begin{tabular}[c]{@{}l@{}}Protests turn violent outside Trump rally in New Mexico\\Protests turn violent outside Trump rally in New Mexico\end{tabular}                                                                                                   & \begin{tabular}[c]{@{}l@{}}japan~(japantoday.com)\\United States (newschannel5.com)\end{tabular}                                         & Different Culture                                                                         & information-crossing      \\ 
\hline
\multicolumn{6}{|l|}{}                                                                                                                                                                                                                                                                                                                                                                                                                                                                                                                                                                                                                                                                                                       \\ 
\hline
\begin{tabular}[c]{@{}l@{}}\textbf{economic }\\\textbf{(Recreation)}\end{tabular} & \begin{tabular}[c]{@{}l@{}}2016-07-13T21:36:00Z\\2016-07-16T19:59:00Z\end{tabular}                       & \begin{tabular}[c]{@{}l@{}}File to seek Gulen's US extradition ready\\Erdogan calls on Barack Obama to extradite Fethullah\end{tabular}                                                                                                                    & \begin{tabular}[c]{@{}l@{}}Azerbaijan~(en.trend.az)\\Armenia~(news.am)\end{tabular}                                                      & \begin{tabular}[c]{@{}l@{}}Similar economic \\Situations (ES)\end{tabular}                & information-not-crossing  \\ 
\cline{2-6}
                                                                                  & \begin{tabular}[c]{@{}l@{}}2018-03-17T20:46:00Z\\2018-03-19T17:14:00Z\\2018-03-22T01:50:00Z\end{tabular} & \begin{tabular}[c]{@{}l@{}}Trump consultants harvested data from 50 million Facebook \\users: reports\\Officials question Facebook's protection of personal data\\Ex-Facebook manager says company was sluggish in \\stopping data harvesting\end{tabular} & \begin{tabular}[c]{@{}l@{}}Pakistan~(geo.tv)\\United States~(union-bulletin.com)\\Kenya~(businessdailyafrica.com)\end{tabular}           & Different ES                                                                              & un-sure                   \\ 
\cline{2-6}
                                                                                  & \begin{tabular}[c]{@{}l@{}}2018-04-03T17:19:00Z\\2018-04-04T18:13:00Z\end{tabular}                       & \begin{tabular}[c]{@{}l@{}}Trump seeks Syria pullout as advisers warn on Islamic State\\White House appears to delay Trump's order for Syrian \\withdrawal\end{tabular}                                                                                    & \begin{tabular}[c]{@{}l@{}}Egypt~(english.ahram.org.eg)\\Iraq (kurdistan24.net)\end{tabular}                                             & Different ES                                                                              & information-crossing      \\ 
\hline
                                                                                  &                                                                                                          &                                                                                                                                                                                                                                                            &                                                                                                                                          &                                                                                           &                           \\ 
\hline
\begin{tabular}[c]{@{}l@{}}\textbf{Political }\\\textbf{(Society)}\end{tabular}   & \begin{tabular}[c]{@{}l@{}}2021-04-07T21:55:00Z\\2021-04-08T06:52:00Z\\2021-04-09T03:22:00Z\end{tabular} & \begin{tabular}[c]{@{}l@{}}Thugs petrol bomb bus as violent riots continue in Belfast\\Thugs petrol bomb bus as violent riots continue in Belfast\\Police Use Water Cannon During Continued Unrest in Belfast\end{tabular}                                 & \begin{tabular}[c]{@{}l@{}}conservatism~(thesun.ie)\\conservatism (thesun.ie)\\centre-right~(theaustralian.com.au)\end{tabular}          & \begin{tabular}[c]{@{}l@{}}Similar Political \\alignment (PA)\end{tabular}                & information-not-crossing  \\ 
\cline{2-6}
                                                                                  & \begin{tabular}[c]{@{}l@{}}2016-01-06T14:51:00Z\\2016-01-08T01:49:00Z\\2016-01-09T01:08:00Z\end{tabular} & \begin{tabular}[c]{@{}l@{}}Iraq offers to mediate in Saudi-Iran crisis stemming from \\cleric's execution\\Iran not seeking tension with Saudi Arabia: Zarif\\Hammond fails to condemn Saudia political executions\end{tabular}                            & \begin{tabular}[c]{@{}l@{}}Neutral (brandonsun.com)\\Conservatism (tehrantimes.com)\\Left-wing (morningstaronline.co.uk)\end{tabular}    & Different PA                                                                              & information-crossing      \\ 
\hline
\multicolumn{6}{|l|}{}                                                                                                                                                                                                                                                                                                                                                                                                                                                                                                                                                                                                                                                                                                       \\ 
\hline
\begin{tabular}[c]{@{}l@{}}\textbf{Linguistic }\\\textbf{(Society)}\end{tabular}  & \begin{tabular}[c]{@{}l@{}}2016-01-15T14:50:00Z\\2016-01-15T17:51:00Z\\2016-01-16T02:20:00Z\end{tabular} & \begin{tabular}[c]{@{}l@{}}Why Amal Clooney doesn't think she's a celebrity\\Amal Clooney Talks About Her New Celebrity Status \\for the First Time\\Amal Clooney Sits Down for First U.S. TV Interview Watch!\end{tabular}                                & \begin{tabular}[c]{@{}l@{}}English (pagesix.com)\\English (vanityfair.com)\\English (usmagazine.com)\end{tabular}                        & \begin{tabular}[c]{@{}l@{}}Similar Publishing \\Language (PL)\end{tabular}                & information-not-crossing  \\ 
\cline{2-6}
                                                                                  & \begin{tabular}[c]{@{}l@{}}2016-01-15T17:38:00Z\\2016-01-15T14:00:00Z\\2016-01-15T16:27:00Z\end{tabular} & \begin{tabular}[c]{@{}l@{}}Friendly no more: Trump, Cruz erupt in bitter fight at \\Republican debate\\The Fight Everyone Wanted to See Finally Happened\\The 6th Republican debate in 100 words (and 4 videos)\end{tabular}                               & \begin{tabular}[c]{@{}l@{}}Spanish (ecodiario.eleconomista.es)\\English (charismanews.com)\\English (scpr.org)\end{tabular}              & Different PL                                                                              & information-crossing      \\ 
\hline
\multicolumn{6}{|l|}{}                                                                                                                                                                                                                                                                                                                                                                                                                                                                                                                                                                                                                                                                                                       \\ 
\hline
\begin{tabular}[c]{@{}l@{}}\textbf{Geographic }\\\textbf{(Society)}\end{tabular}  & \begin{tabular}[c]{@{}l@{}}2021-04-28T11:48:00Z\\2021-04-28T22:52:00Z\end{tabular}                       & \begin{tabular}[c]{@{}l@{}}Lawmaker Says Schools Must Teach The 'Good Of \\Slavery' (Video)\\Backlash on Louisiana lawmaker grows following his \\comments about slavery\end{tabular}                                                                      & \begin{tabular}[c]{@{}l@{}}United States~(patheos.com)\\United States (theadvertiser.com)\end{tabular}                                   & \begin{tabular}[c]{@{}l@{}}Publishers' \\headquarters in \\same country\end{tabular}      & information-not-crossing  \\ 
\cline{2-6}
                                                                                  & \begin{tabular}[c]{@{}l@{}}2016-06-10T11:21:00Z\\2016-06-10T14:29:00Z\\2016-06-10T15:18:00Z\end{tabular} & \begin{tabular}[c]{@{}l@{}}Queen's dedication praised at 90th\\Queen's dedication praised at 90th\\Queen's dedication praised at 90th\end{tabular}                                                                                                         & \begin{tabular}[c]{@{}l@{}}England~(haverhillecho.co.uk)\\newsletter.co.uk (United Kingdom)\\Australia (adelaidenow.com.au)\end{tabular} & \begin{tabular}[c]{@{}l@{}}Publishers' \\headquarters in \\different country\end{tabular} & information-crossing      \\ 
\hline
\multicolumn{1}{l}{}                                                              & \multicolumn{1}{l}{}                                                                                     & \multicolumn{1}{l}{}                                                                                                                                                                                                                                       & \multicolumn{1}{l}{}                                                                                                                     & \multicolumn{1}{l}{}                                                                      & \multicolumn{1}{l}{}     
\end{tabular}}
\end{table*}
\FloatBarrier

\section{Materials and methods}\label{sec:MethodsforAutomaticBarrierProfiling}
In this section, we present an analysis of sentiments across different barriers followed by the properties of commonsense inferences knowledge, classification baselines, evaluation metric, and experimental results comparing simple (LR, SVM, DT, RF, kNN), deep learning (LSTM), and transformers (BERT) for the barrier classification task.

\subsection{Analysis of sentiments}
\begin{figure*}
\centering
\includegraphics[keepaspectratio=true,scale=0.50]{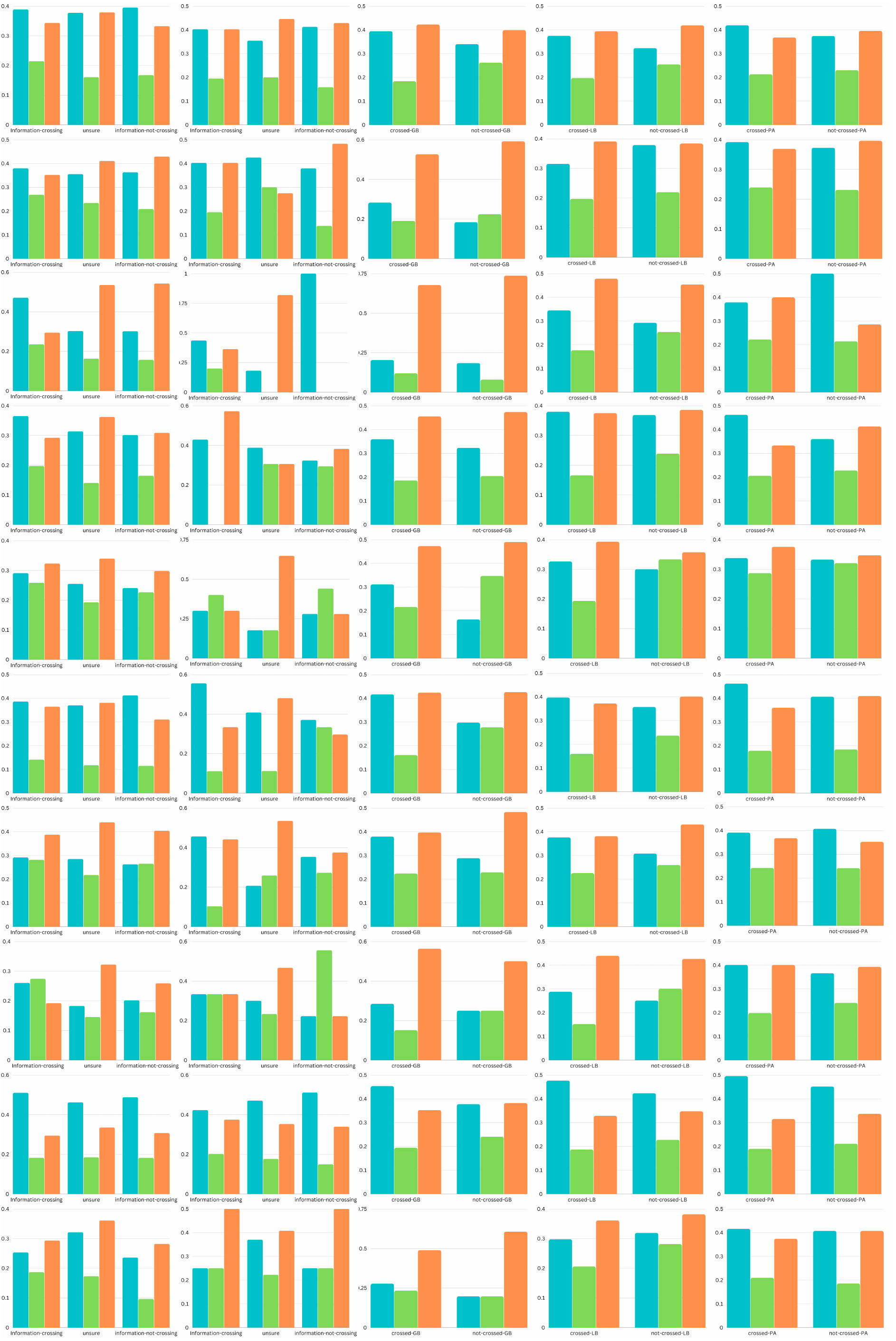}
\caption{\bf The bar charts present the sentiments of the headlines for each category. It shows the comparison of sentiments (positive, neutral, and negative) for all ten categories. Each column represents (from the top to bottom) the ten categories (business, computers, games, health, home, recreation, science, shopping, society, and sports), and each row represents (from the left to right) the five barriers (cultural, economic, geographic, linguistic, and political).}
\label{fig:sentimentComparision}
\end{figure*}

We use the Vader rule-based model to obtain the emotional and sentiment polarity of the news headlines to analyze the variation of sentiments across the different categories of the different barriers. Vader provides a polarity range of the news headlines in the interval from -1 to +1. The −1 value represents a negative polarity and +1 indicates a positive polarity \cite{ref:martin2021suspicious}. The bar charts illustrate the differences in sentiments across binary and ternary classes in each category of the five barriers (see Figure \ref{fig:sentimentComparision}). For each instance, we have one of the three sentiments: positive, neutral, and negative. 

For the binary class classification of the political, linguistic, and geographical barrier, the headlines that have been labeled as crossing the barrier have the following sentimental differences: the categories business, home, health, recreation, science, shopping, and society have more instances of negative sentiments than positive and neutral with the considerable differences of (8\%, 1\%, 10\%, 5\%, 8\%, 5\%, 5\%), (6\%, 5\%, 2\%, 5\%, 8\%, 5\%, 5\%), and (7\%, 12\%, 4\%, 12\%, 9\%, 3\%, 5\%) respectively; The game category of news headlines with annotation of crossing the political barrier has 12\%, 6\%, 1\% more instances of positive sentiments for the political, linguistic, and geographical barrier; The news headlines that have been labeled as not crossing the political barrier have the following differences: the categories computers, health, recreation, science, society, and sports have more instances of positive sentiments than neutral and negatives sentiments with the considerable differences of (5\%, 8\%, 5\%, 5\%, 2\%, 5\%), (7\%, 2\%, 5\%, 5\%, 2\%, 3\%), and (4\%, 2\%, 1\%, 8\%, 3\%, 5\%) respectively; The game category has 10, 1\% and 3\%  more instances respectively with negative instances than other classes; With regard to the ternary classification of the cultural barrier, the news headlines that have been labeled as crossing the barrier, have the following sentimental differences: the categories games, health, shopping, and society have more instances of negative sentiments than other classes with the considerable differences of 12\%, 5\%, 6\%, and 2\% respectively. The news headlines that have been labeled as not crossing the cultural barrier have the following differences: the categories business, computers, health, recreation, science, shopping, society, and sports have more instances of positive sentiments than other classes with considerable differences of 5\%, 6\%, 5\%, 5\%, 10\%, 3\%, and 8\% respectively. The news headlines that have been labeled as unsure, have the following differences: the categories computers, and shopping, have more instances of positive sentiments than other classes with considerable differences of 6\%, and 5\% respectively whereas game has 20\% more instances of negative sentiments; With regard to the ternary classification of economic barrier, the headlines that have been labeled as crossing the barrier, have the following sentimental differences: the categories business, home, recreation, science, shopping, and sports have more instances of negative sentiments than other classes with the considerable differences of 5\%, 7\%, 8\%, 5\%, 2\%, and 14\% respectively. The news headlines that have been labeled as not crossing the economic barrier have the following differences: games, home, recreation, science, and shopping have more instances of positive sentiments than other classes with considerable differences of 18\%, 16\%, 7\%, 5\%, and 8\% respectively. The headlines that have been labeled as unsure, have the following differences: business and games have 5\% and 22\% respectively more instances of negative sentiments, computers, and sports have 8\%, and 10\% respectively more instances of positive sentiments and recreation and shopping have 12\% and 13\% more instance of neutral sentiment respectively.

Overall with regard to the binary class classification for the political, linguistic, and geographical barriers, we see that the news headlines that are labeled as crossing the barrier, have more instances of negative sentiments, whereas the news headlines that are labeled as not crossing the barrier, have more instances with positive sentiments. With regard to the ternary class classification for the economic and cultural barrier, we see that the news headlines that are labeled as crossing the barrier have more instances of negative sentiments, whereas the news headlines that are labeled as not crossing the barrier, have more instances with positive sentiments. However, in case of the news headlines that are labeled as unsure, have more instances of negative sentiments for the economic barrier and positive sentiments for the cultural barrier.

\begin{figure*}
\centering
        \includegraphics[width=0.95\textwidth]{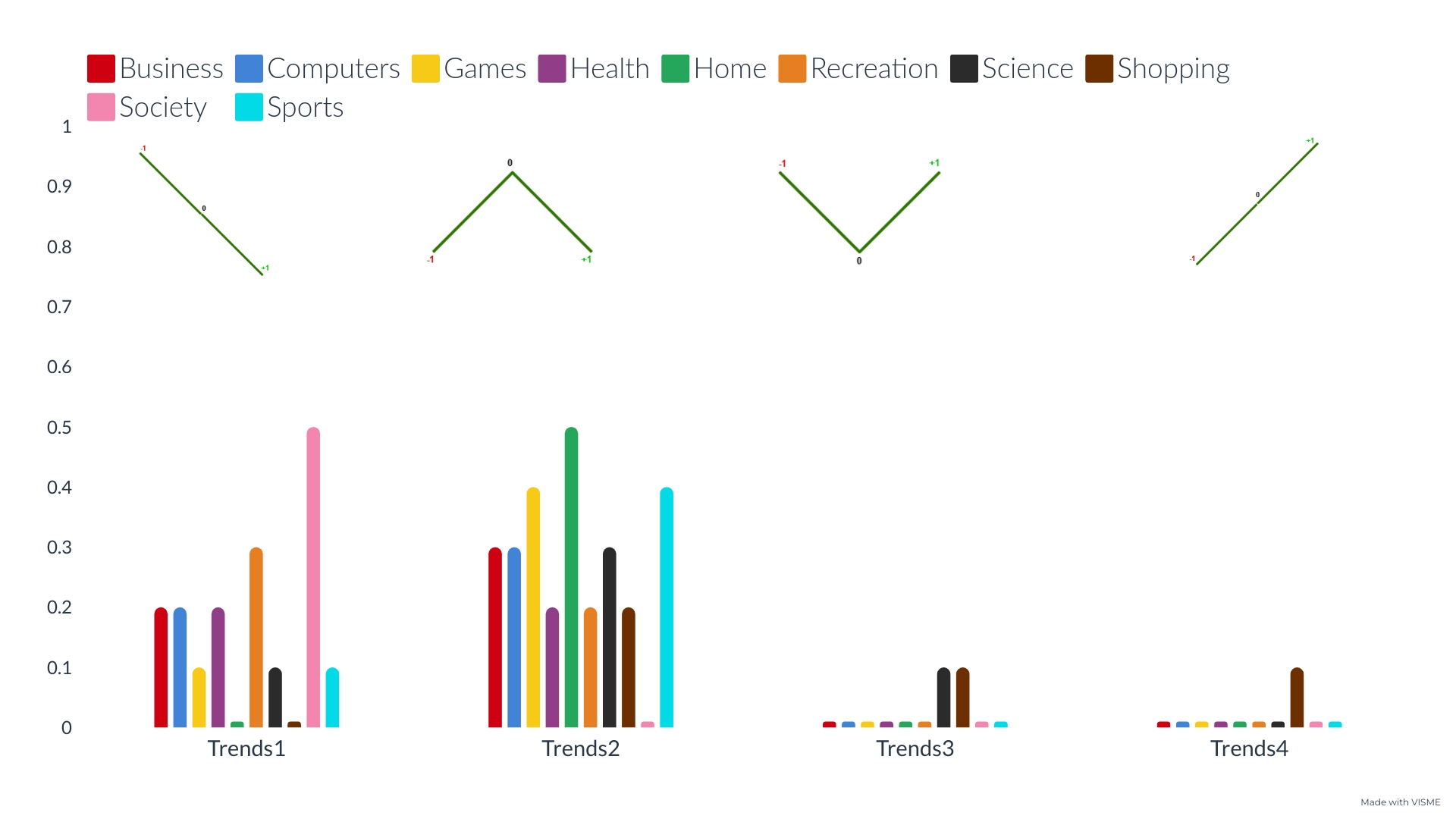}
        \hspace{2cm}
        \includegraphics[width=0.95\textwidth]{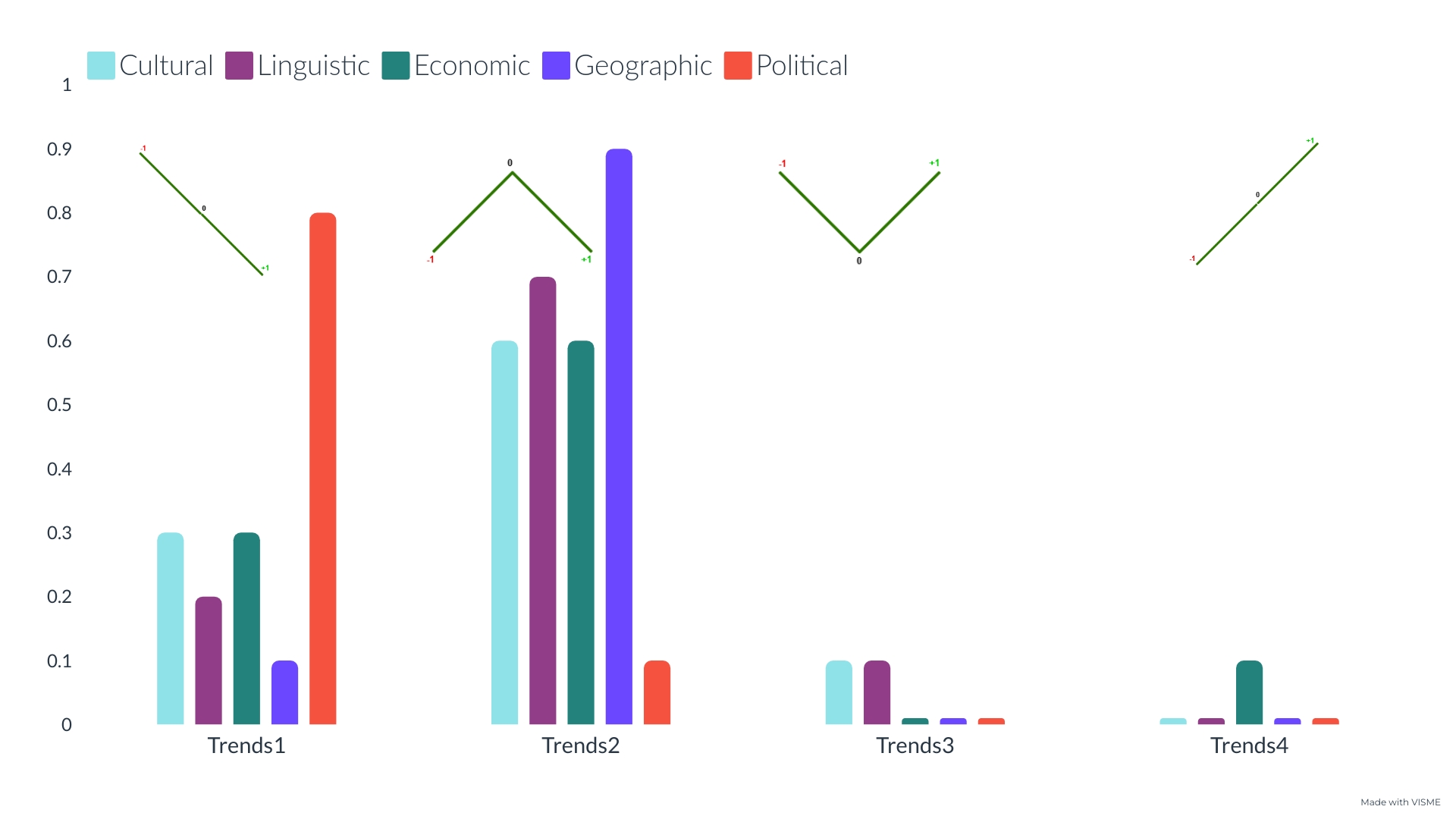}
        \caption{\bf The bar charts present the distribution of different possible trends of sentiments across different categories and barriers (from left to right). The sentimental trends vary in four different types (see on the x-axis): trend1, and trend4 represent decrement and increment respectively in the percentage of news articles (see on the y-axis) with negative sentiment to neutral and then to positive: trend2, and trend3 represent decrement and increment respectively in the percentage of news articles with neutral sentiments than positive and negative sentiments.}
        \label{fig:sentimentalTrends}
\end{figure*}
\FloatBarrier

The bar charts present the distribution of different possible trends of sentiments across different categories, and barriers (see Figure \ref{fig:sentimentalTrends}). We analyzed the sentimental trends and found that among other possible trends, these four trends are covering more than 95\% of data. The first trend shows that the number of positive instances are higher than neutral instances and then these both are higher than negative instances. The fourth trend is reverse of it. The second trend shows that the number of neutral instances are higher than positive and negative, and, negative and positive instances are approximately equal to each other. The first bar chart shows that more than 30\% of news of society and recreation categories consist of news headlines with positive sentiment whereas more than 30\% of news of games, home, and sports categories consist of news headlines with neutral sentiments. The second line graph shows that 80\% of news headlines belonging to the political barrier have negative sentiment whereas ninety 90\% of news headlines belonging to the geographical barrier have neutral sentiments.

The results suggest us the following conclusions for the \textbf{Q1}: 1) the political barrier has been crossed by the news with positive sentiments and reverses for the other four barriers (linguistic, geographic, cultural, and economic). The news with negative sentiments has not been crossing the political barrier and has been crossing the linguistic, geographic, cultural, and economic barrier; 2) the variations in the sentiments across binary and ternary class classifications of the different categories of news and the barriers suggest that we should take sentiment score as a feature in barrier classification. \cite{ref:alonso2021sentiment} have considered sentiments of news for fake news detection based on the fact that sentiment is a complementary element to fake news.

\subsection{Commonsense knowledge}
We use a common sense knowledge resource COMET atomic through an inference toolkit called \emph{kogito} to generate commonsense inferences in a given situation by assessing their intentions and behaviors. This toolkit provides the interface to interact with natural language generation models that can be used to infer the commonsense from a textual input \cite{ref:hwang2021comet},\cite{ref:Ismayilzada2022kogito}. These models consist of triplets of head entity, relation, and tail entity. We present an example illustrating the results of the commonsense of different relations about three news headlines (taken from the table \ref{tab:annotationExamples}). The first headline ("Uproar at Republican convention as anti-Trump delegates revolt") has six relations such as \emph{react, need, intend, want, isFilledBy, and react}. To convert the common sense knowledge into a meaningful text we consider each tuple consisting of the relation, and tail as a sentence and then concatenate them. To make the tuples as sentences, we change the relation to the past form such as reacted angry, needed to make a speech, intended to protest, wanted to protest, isFilledBy uproar at the republican convention, and reacted upset.  

\begin{figure*}
\centering
\includegraphics[keepaspectratio=true,scale=0.80]{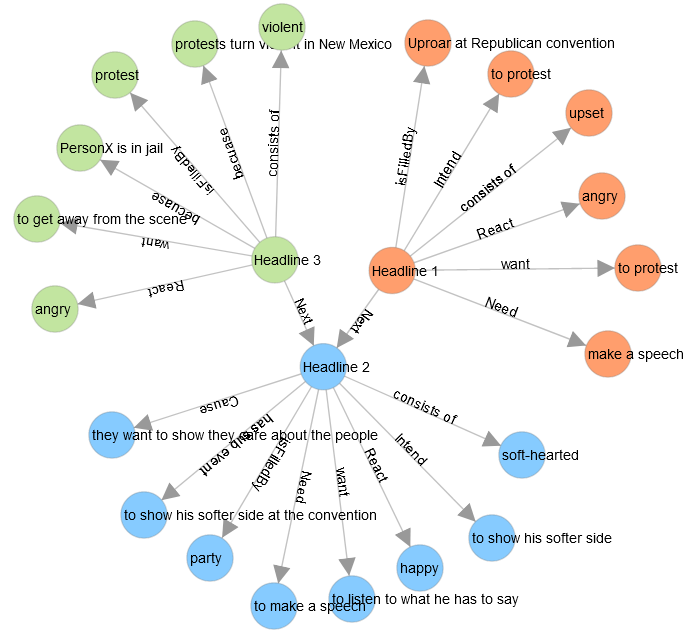}
\caption{\bf The network diagram presents an example of headlines with the common sense knowledge. Headline1 is "Uproar at Republican convention as anti-Trump delegates revolt", Headline2 is "Trump aims to show his softer side at Cleveland convention", and Headline 3 is "Protests turn violent outside Trump rally in New Mexico" (These headlines belong to the cultural barrier (see Table \ref{tab:annotationExamples})).}
\label{fig:networkDiag}
\end{figure*}
\FloatBarrier

\begin{figure*}
\centering
        \includegraphics[width=0.30\textwidth]{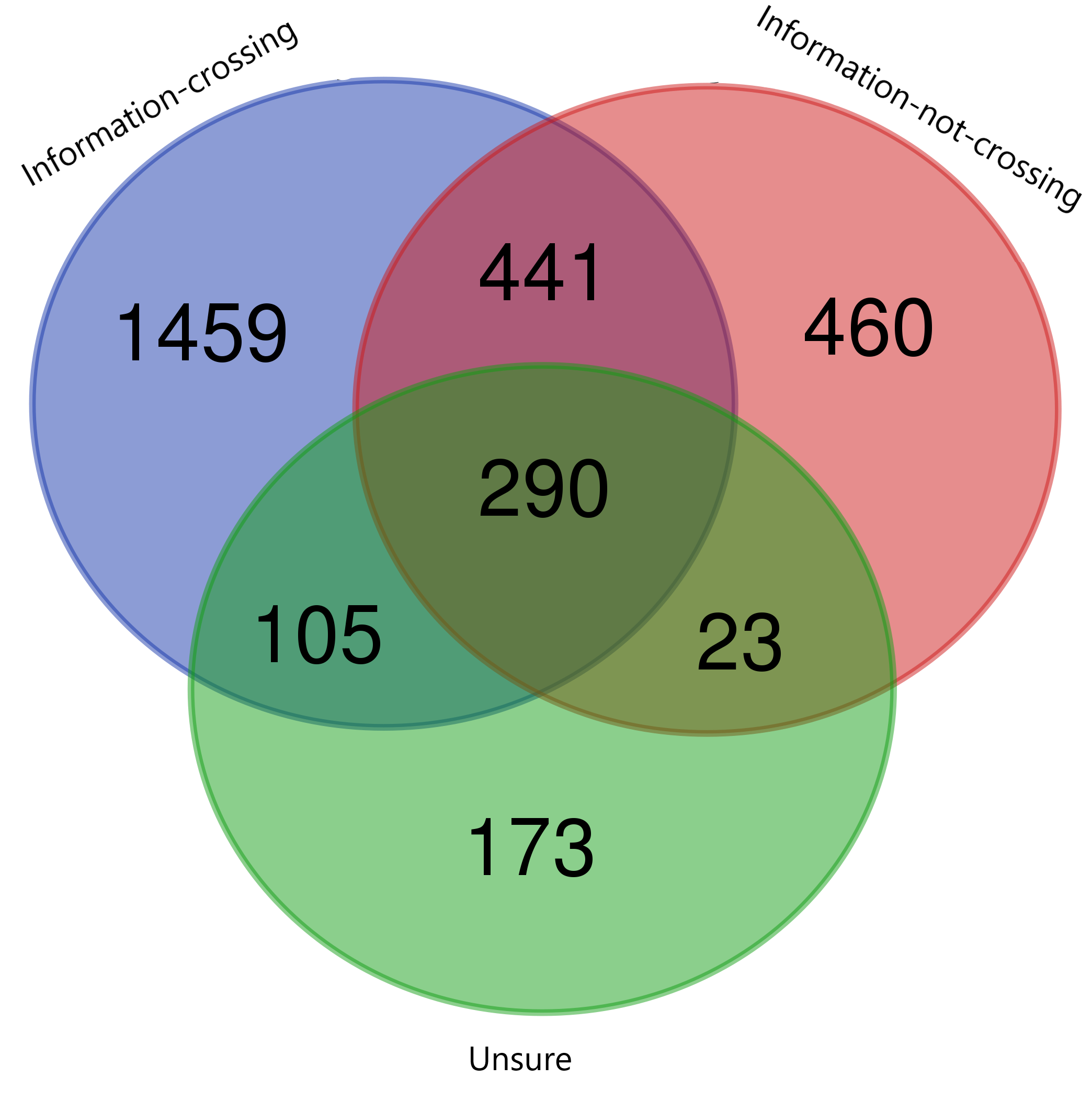}
        \hspace{2cm}
        \includegraphics[width=0.40\textwidth]{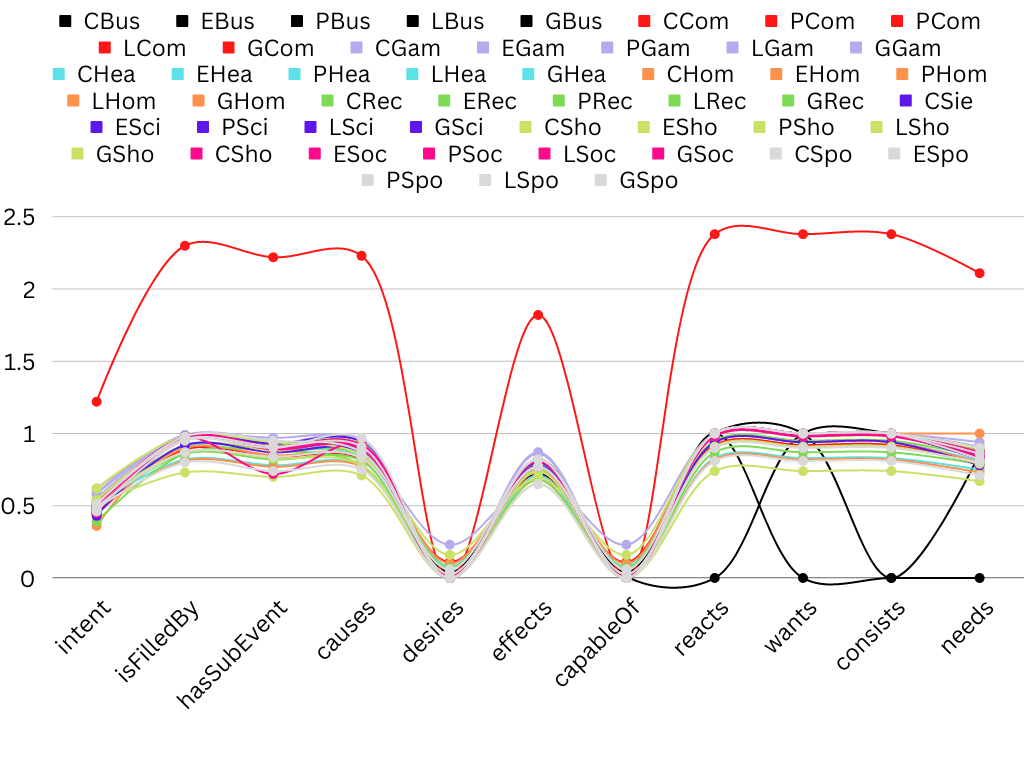}
        \caption{\bf The Venn diagrams show the intersection between inferences across ternary classes of the cultural barrier. The line graph shows the frequency of all the inferences to all categories of all the barriers.}
        \label{fig:venn_result-inferences}
\end{figure*}

The purpose of using the semantic knowledge in form of commonsense knowledge was to improve text classification, we analyzed the associated inferences to all the barriers. We present an example to illustrate the comparison. We choose the cultural barrier and to perform a comparison between the categories, we select the category society. The results of the intersection between the inferences belonging to three different classes (information crossing, information not crossing, and unsure) have been shown in Figure \ref{fig:venn_result-inferences}. There are 290 inferences that are common among all three classes and there are 441, 105, and 23 inferences that are common between class one and two, class two and three, and class one and three respectively. The most important fact is that there are 1459, 173, 460 unique inferences for the class one, two and three respectively that can be useful for the classification in this ternary class classification. 
\newline

\FloatBarrier

To answer the \textbf{Q2} the line graphs in Figure \ref{fig:venn_result-inferences} present the statistics of inferences across the ten different categories of the five barriers. Since the main purpose in this figure is to analyze the statistics of the inferences across the categories, we keep the same color for a category across the five barriers. The x axis shows the name of all the inferences and the y axis shows the average number of inferences per a news headline in each category. The categories that have significant differences are computer, and business. The average inferences are significantly high in the computer category than other all categories. Each news headline contains 1.5\% inference type "intent"  and consists of "isFilledBy", "hasSubEvent", "Causes", "reacts", "wants","consists", and "needs" inferences types with average of approximately 2.5. The existence of the inference type is almost 0\% per news headline for "desires" and "capableOf". For the category business, the existence of a few types of inferences are equal to zero such as "reacts", "wants", "consists", and "needs". Otherwise, the average of the existence of all the inferences per news headline is approximately equal for all the categories of all the barriers. 
\newline
This analysis makes us understand the distribution of different inferences across different categories, as well as associated semantic knowledge per news headlines. Since different features such as sentiments, and semantic knowledge possess different discriminative capabilities in classification \cite{ref:nassirtoussi2015text, ref:zhai2011exploiting}, we use them as classification of news spreading barrier.

\subsection{Evaluation methodology and baselines}
\begin{figure*}
            \centering
            \includegraphics[keepaspectratio=true,scale=0.24]{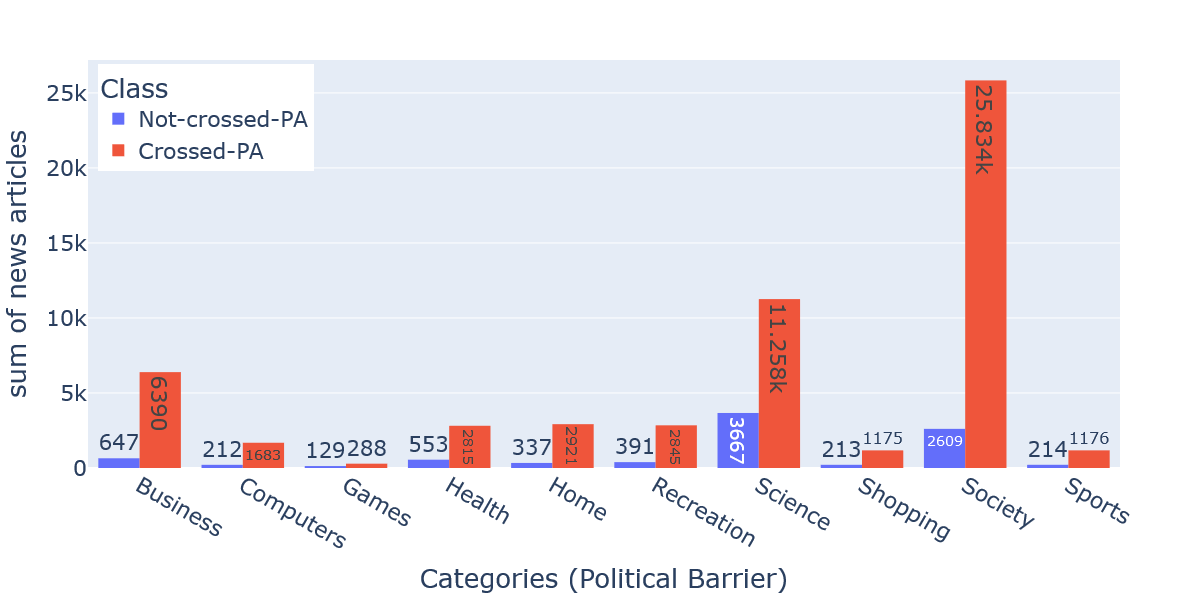}
            \includegraphics[keepaspectratio=true,scale=0.24]{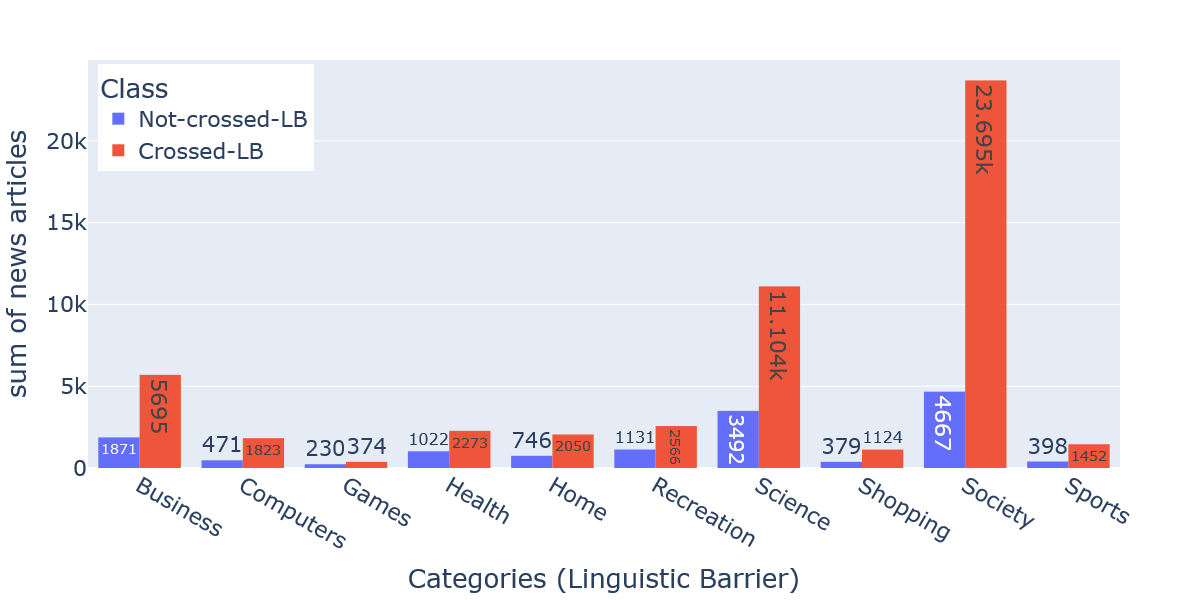}
            \includegraphics[keepaspectratio=true,scale=0.24]{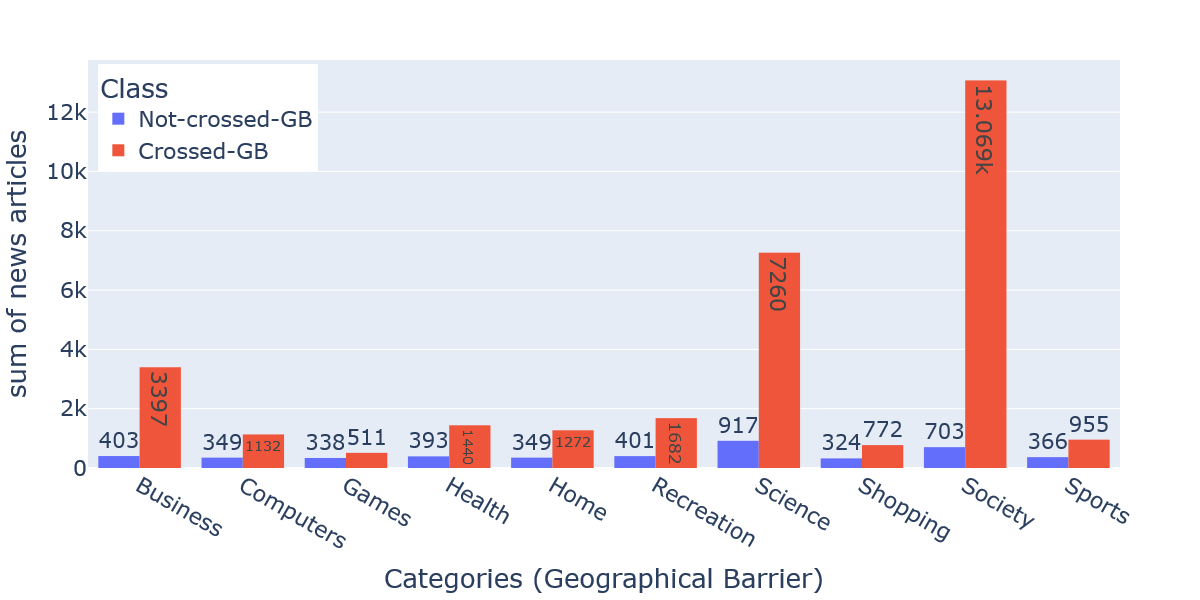}
            \caption{\bf This bar chart shows the class distribution for the political, linguistic, and geographical barriers (from left to right). The bar with blue color shows the distribution for the class "Information-crossing" a barrier whereas the bar with red color shows the distribution for the class "Information-not-crossing" a barrier. Each of the three-bar charts presents the class distribution for all ten categories.}
            \label{fig:barriDist}
\end{figure*}

\begin{figure*}
            \centering
            \includegraphics[keepaspectratio=true,scale=0.28]{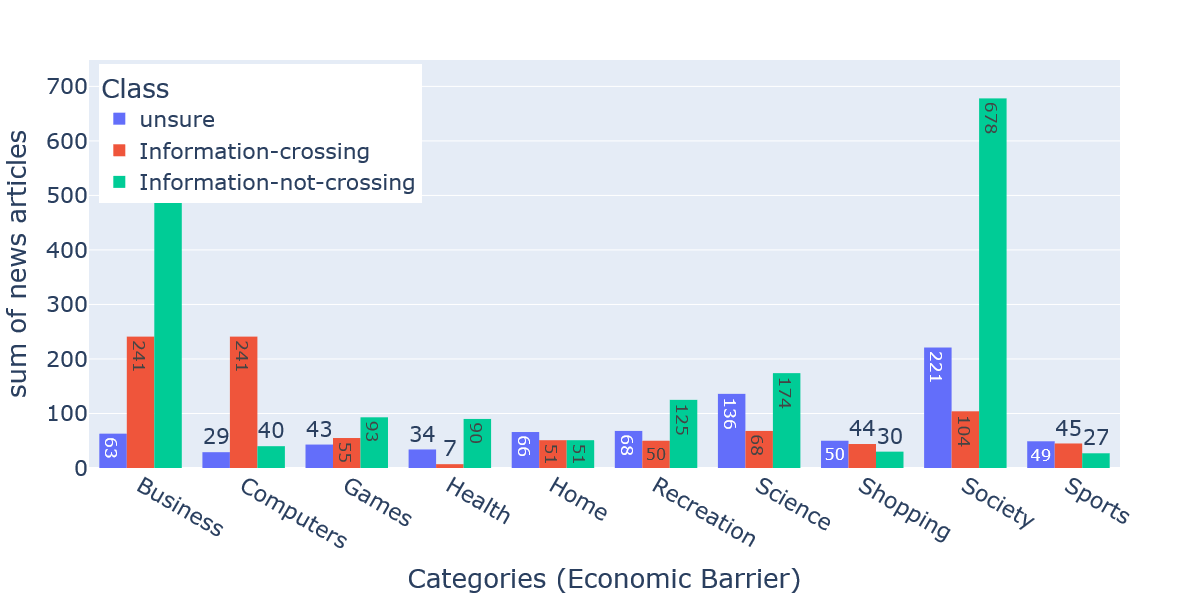}
            \includegraphics[keepaspectratio=true,scale=0.28]{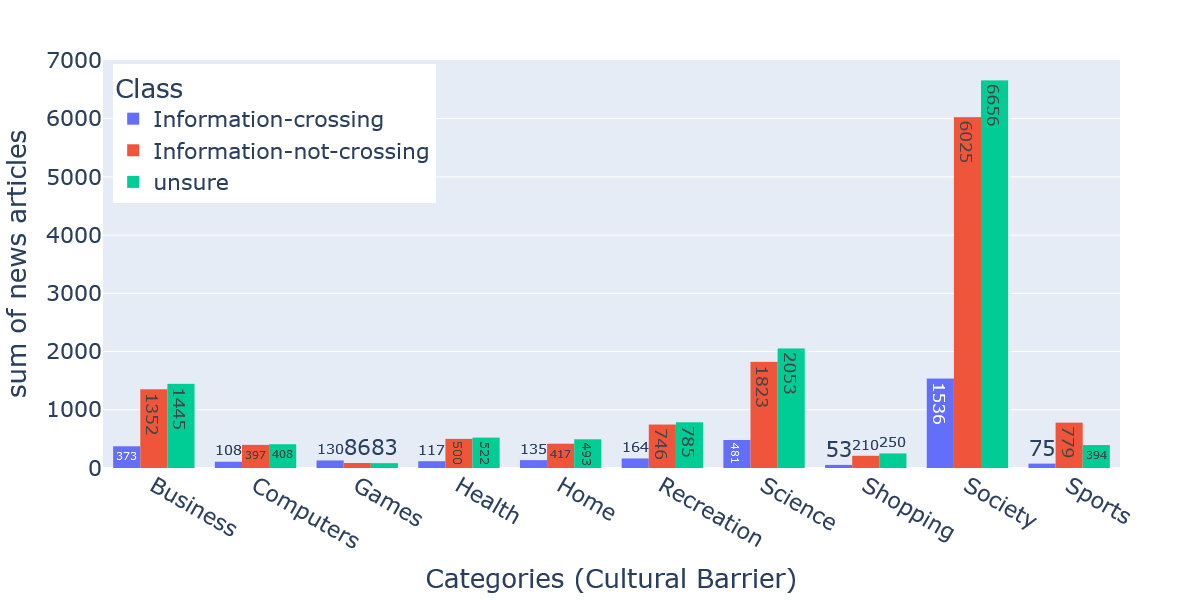}
            \caption{\bf This bar chart shows the class distribution for the economic, and cultural barriers (from left to right). The bar with red color shows the distribution for the class "Information-not-crossing" whereas the bar with green color shows the distribution for the class "Unsure" a barrier. The bar with blue color shows the distribution for the class "Information-crossing". Each of the two bar charts presents the class distribution for all ten categories.}
            \label{fig:ecoculDist}
\end{figure*}

We used Scikit-learn implementation of classical and deep learning models considering the following parameters, which are usually the default: hidden layers = 3, hidden units = 64, no. of epochs = 10, batch size = 64, and dropout = 0.001. For the training process of political, geographical, and linguistic barrier, we used Adam as the optimizer, binary cross-entropy as the loss function, and sigmoid as the activation function. For economic and cultural barriers, we used Adam as the optimizer, categorical cross-entropy as the loss function, and SoftMax as the activation function. The data about each barrier is split into train-set and test-set with a ratio of 80\%-20\%. To maintain the class proportion in the train and test set, we use stratified sampling (see Figures \ref{fig:barriDist}, \ref{fig:ecoculDist}), meaning that the training and testing set must have equal proportion of all the classes. For instance, if in case of business category of culture barrier, there are total 100 news headlines where 40 instances have label "information crossing", 40 instances have label "information not crossing", and 20 instances have label "Unsure". And, we suppose we split our train and test set in the ratio of 80:20. Then, the train set and test set must have 20, 20, and 10 instances of each label ("information crossing", "information crossing", and "Unsure") respectively.


For the comparison with the proposed common sense inferences and semantic knowledge, we evaluated the barrier classification task using the news headlines text only. After performing the preprocessing step such as lower case conversion and stop words removal, we adopted the term frequency (TF) and inverted document frequency (IDF) methods to represent the bag of words of each news article \cite{ref:yazdani2017sentiment}. For the barrier classification task, the experiments were conducted by utilizing three different types of machine learning algorithms: 1) classical machine learning algorithms including Logistic Regression (LR), Naive Bayes (NB), Support Vector Classifier (SVC), k-nearest Neighbor (kNN), and Decision Tree (DT): The performance of LR for the text classification problems is same as of the SVM algorithm \cite{ref:shah2020comparative}. SVMs use kernel functions to find separating hyper-planes in high-dimensional spaces \cite{ref:colas2006comparison}. SVM is difficult to interpret and there have to be many parameters that need to be set for performing the classification and one parameter that performs well in one task might perform poorly in other\cite{ref:shah2020comparative}. Therefore many information retrieval systems use decision trees and naive bayes. However, these models lack accuracy \cite{ref:kowsari2017hdltex, ref:kamath2018comparative}. 2) LSTM (Long-Sort-term Memory): With the emergence of deep learning algorithms, the accuracy of text categorization has been greatly improved. Convolutional neural networks (CNN) and long short-term memory networks (LSTM) are widely used \cite{ref:luan2019research, ref:yu2020attention, ref:luan2019research, ref:kamath2018comparative, ref:wang2017comparisons}. 3) State-of-the-art pre-training language model BERT (Bidirectional Encoder Representations from Transformers): It is trained on a large network with a large amount of unlabeled data and adopts a fine-tuning approach that requires almost no specific architecture for each end task and has achieved great success in a couple of NLP tasks, such as natural language inference, and text classification \cite{ref:yu2019improving, ref:jin2020bert, ref:gonzalez2020comparing}.

\subsection{Evaluation metric}\label{subsec:ExoSetyo}
To evaluate the performance of binary and multi-class barrier classification models, F1-score is used as an evaluation measure.
\begin{itemize}
  \item \textbf{F1-Score:} It combines the precision and recall of a classifier into a single metric by taking their harmonic mean.
  It is defined as:
                     \[ F_{1} =  \frac {2(Precision * Recall)}{Precision + Recall} \] 
\end{itemize}

\subsection{Comparative analysis of the ten categories} \label{subsec:comAnaTenCates}
We compare the results of all ten news categories based on the evaluation metric F1-score. Since the results of LR among the five (LR, SVC, NB, DT, and kNN) classical machine learning algorithms were higher in all categories, we exclude the others. Table \ref{tab:resultsTable} compares the results of LR, LSTM, and BERT with our proposed approach that is based on commonsense-based semantic knowledge and sentiment. The words PM-LSTM (proposed model LSTM) and PM-BERT (proposed model BERT) mean the usage of LSTM and BERT utilizing our approach with the inferences-based semantic knowledge and sentiments. For the cultural barrier, F1-scores using BERT or LSTM with commonsense-based semantic knowledge and sentiment are higher than LR, LSTM, and BERT for business, computers, games, health, home, recreation, science, shopping, society, and sports 
(with an improvement of 0.02, 0.05, 0.01, 0.09, 0.11, 0.14, 0.09, 0.12, 0.06, and 0.03 respectively.); For the economic barrier, F1-scores are higher than LR, LSTM, and BERT for business, computers, health, home, and sports (with an improvement of 0.06, 0.03, 0.1, 0.14, and 0.01 respectively.); For the Political barrier, F1-scores are higher than LR, LSTM, and BERT for business, computers, games, health, home, recreation, science, shopping, and society (with an improvement of 0.12, 0.28, 0.08, 0.13, 0.21, 0.07, 0.1, 0.14,  and 0.24 respectively.); For the Linguistic barrier, F1-scores are higher than LR, LSTM, and BERT for science, and society (with an improvement of 0.26, and 0.24 respectively.); For the geographical barrier, F1-scores are higher than LR, LSTM, and BERT for business, computers, health, home, recreation, and society (with the improvement of 0.03, 0.44, 0.25, 0.17, 0.26, and 0.28 respectively.); 
\newline
\newline
The results presented in Table \ref{tab:resultsTable} indicate that the best results for each barrier are obtained by PM-BERT and BERT, however closely followed by the PM-LSTM. The LR performs a little less compare to the other algorithms were tested. Moreover, it can be seen that the obtained F1-score vary significantly across different categories. While the obtained F1-score is very high for the two categories (health, and society) of geographical barrier, for the three categories (business, shopping, and science) of political barrier, and a quite good score is obtained for recreation category of cultural barrier, games category of economic barrier, and computers category of linguistic barrier, the score is low comparatively for the other categories of the different barriers. 
\newline
\newline
The best results obtained for the task of classifying the barriers for the ten different categories is a direct consequence of the class distribution and sentiments of the classes to some extent. As far as the results are concerned for all the barriers, we see that the highest F1-score is produced for the health (0.97), and society (0.97) categories of geographical barrier, recreation (0.66) category of the cultural barrier, games (0.72) category of the economic barrier, computers (0.97) category of the linguistic barrier, and business (0.97), shopping (0.97), and science (0.97) categories of the political barrier. The F1-score for society category of geographic and business category of political is as high as equal to 0.97. An obvious reason for this is the fact that the data is heavily imbalanced with 95\% and 91\% instances of majority classes. However, still both are showing the improvements. This can be due to a slight variation of sentiments across its binary classes. For the health category of geographical barrier, and shopping and science category of political barrier, the class distribution is not much imbalanced (78\%, 85\%, and 75\% instances of majority class) but F1-score is really high that means PM-BERT is best suited for these categories. Regarding its best results, it might be possible that sentiments across these binary classes have variations such as label "Crossed-GB" have more positive and less negative instances than "Not-crossed-GB" and vice versa. Similarly, the shopping and science category of political barrier consist of more news headlines with negative sentiments for label "Crossed-PB" and vice versa for the label "Not-crossed-PB". PM-BERT has proved to be best suited for the classification of computers and games categories of linguistic and economic barriers. We see that the data is quite balanced for computers category of linguistic (75\% and 25\% instances of "Crossed-LB", "Not-crossed-LB" respectively) and games category of economic barrier (29\%, 49\%, 22\% instances of "information-crossing", "information-not-crossing", and "unsure" classes respectively). Looking into the sentiments of each class of computers category (see Figure \ref{fig:sentimentComparision}), we observe that one has more news headlines with positive headlines and less negative news headlines and vice versa. However, for the games category of economic barrier, sentiments are varying across all three classes: 12\%, 25\%, and 50\% news headlines with positive, neutral, and negative sentiments respectively for the label "Information-crossing". The label "Unsure" does not have news headlines with neutral sentiments whereas all the news headlines labeled as "Information-not-crossing" has only positive sentiments. For the recreation category of cultural barrier, although the distribution of positive, neutral, and negative sentiments across all three barriers is almost equal as well as the class distribution is balanced (10\%, 44\%, and 46\% instances for "Information-crossing", "Information-not-crossing", and "Unsure" respectively), the PM-BERT performs really well (0.66 F1-score).

\begin{figure}
\centering
\includegraphics[keepaspectratio=true,scale=0.18]{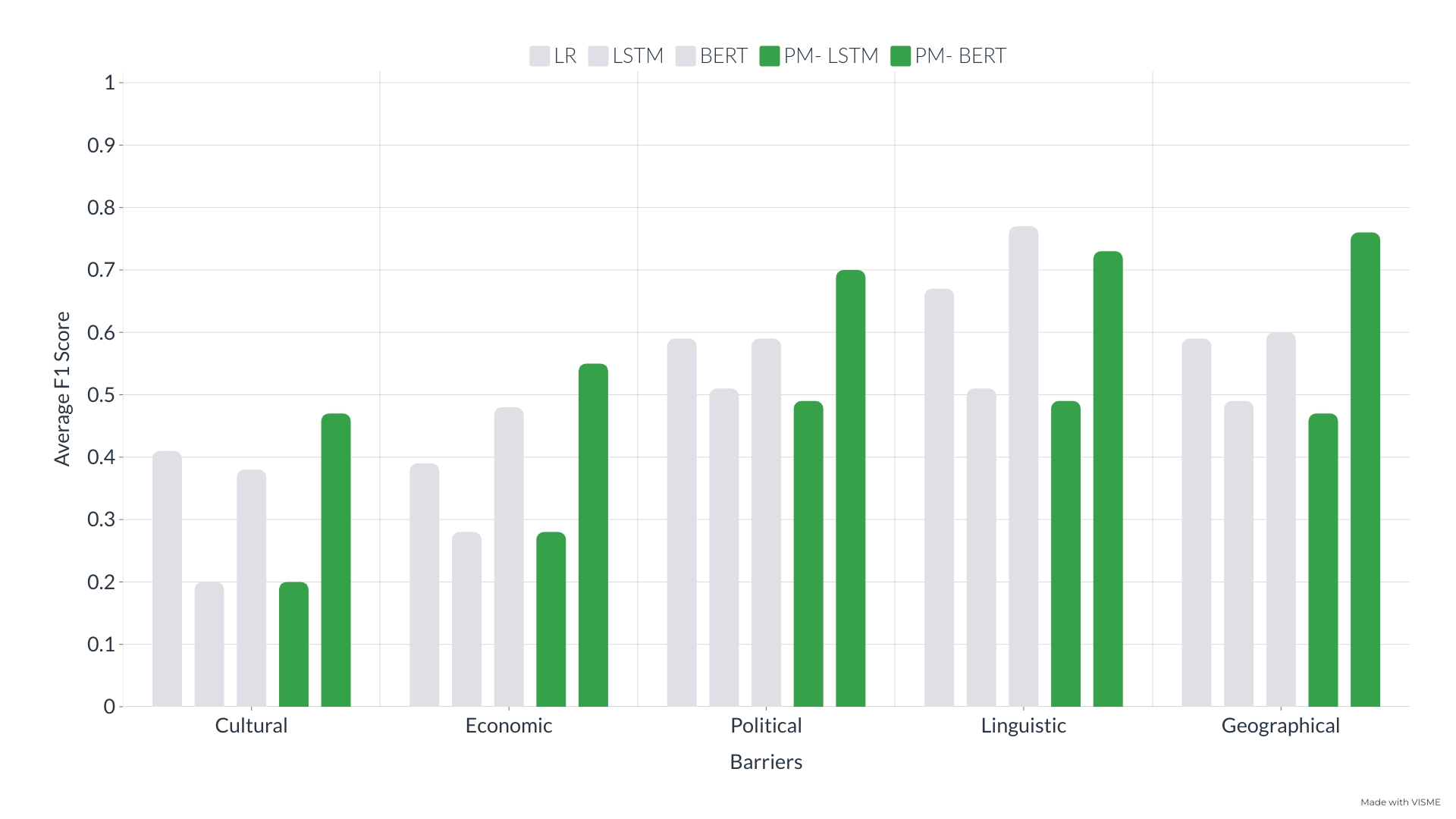}
\caption{It presents the bars of two colors for each barrier. The green bars show the average F1-scores of all the ten categories for LSTM and BERT using commonsense-based semantic knowledge and sentiments. The gray bars show the average F1-scores of all the ten categories for LR, LSTM, and BERT using only headline text. The x-axis shows the groups of bars for all five barriers whereas the y-axis shows the average F1-score.}
\label{fig:H-Algorithms}
\end{figure}

\subsection{Comparative analysis of three types of algorithms} \label{subsec:comAnaThreAlgos}
After discussing the results of all ten news categories, we compare all the five different types of barriers using the average F1-scores of all ten categories. Figure \ref{fig:H-Algorithms} presents the comparison of the average F1-score of all the categories. The highest average F1-score for the cultural barrier is obtained using PM-BERT (0.47) and BERT (0.38) whereas this score was low using PM-LSTM (0.22). For the economic, political, and geographical barriers, the average F1-score obtained using PM-BERT (0.55, 0.70, and 0.76 respectively) followed by a little lower average F1-score using BERT (0.48, 0.59, and 0.60 respectively) and then PM-LSTM (0.28, 0.49, and 0.57 respectively). 
\newline
\newline
With regards to the linguistic barrier, the highest average F1-score achieved using BERT instead of PM-BERT or PM-LSTM which is quite interesting and questionable. Instead of average, we look for the F1-score of all the ten categories of this barrier. The obtained F1-score for the eight categories (business, computers, health, home, science, shopping, sports and society) using PM-BERT is high then BERT (using PM-BERT : 0.80, 0.97, 0.82, 0.83, 0.78, 0.67, 0.73, 0.79 respectively; using BERT : 0.74, 0.75, 0.79, 0.81, 0.78, 0.63, 0.73, 0.76 respectively). However, the F1-score for the game and recreation categories of the linguistic barrier using PM-BERT (0.47, and 0.48 respectively) was considerably lower than BERT (0.85, and 0.81 respectively). On the other hand, the obtained average F1-score is better for four barriers including economic, political, linguistic, and geographical than cultural barrier, that is just under 0.5. However, for the other barriers, it is well over 0.50 and extends to near 0.80. To answer the \textbf{Q3}, we can say that our proposed methods (LSTM and BERT with semantic knowledge) outperform for the four cultural, economic, political and geographical barrier.

\subsection{Analysis and discussion}\label{subsec:AnalysisAndDiscussion}
Experiments of the novel approach on the ten different kinds of news and for the five different barriers have brought some insights regarding information spreading. In order to support the hypothesis, we have set three research questions. To answer the first research question (Do the sentiments of the headlines of different topics vary across the different barriers?), 
we compare the sentiments of the headlines for all the categories across the five barriers performing sentiments analysis at the granularity of ten negative and ten positive points as well as at overall negative, positive, and neutral sentiment  (see Figures \ref{fig:sentimentComparision}, \ref{fig:sentimentalTrends}). The comparative analysis indicates that the political barrier has been crossed by the news with positive sentiments whereas for the other four barriers (linguistic, geographical, cultural, and economic) news with positive sentiments are not crossing the barriers. With regard to the binary class classification for the political, linguistic, and geographical barrier, we see that the news headlines that are labeled as crossing the barrier, have more instances of negative sentiments, whereas the news headlines that are labeled as not crossing the barrier, have more instances with positive sentiments. With regard to the ternary class classification for the economic and cultural barrier, we see that the news headlines that are labeled as crossing the barrier have more instances of negative sentiments, whereas the news headlines that are labeled as not crossing the barrier, have more instances with positive sentiments. To answer the second research question (What are the properties (statistics and ratio) of the commonsense knowledge relations in news headlines to different topics?), we find the intersection between the inferences belonging to different barriers and categories (see Figure \ref{fig:networkDiag}, \ref{fig:venn_result-inferences}). The results suggest that although inferences are being shared among the classes, there are some unique inferences for each class. Similarly, the same fact exists between the different categories. Therefore it might be possible that it will help in improving the classification results. The results of annotation show that there are variations in class distributions across different categories. Therefore, we use stratified sampling to maintain the class proportion in the train and test sets (see Figures \ref{fig:barriDist}, \ref{fig:ecoculDist}). To answer our third research question (Which classification methods (classical or deep learning methods) yield the best performance to barrier classification task?), We perform classification with classical machine learning methods including Logistic Regression (LR), Naive Bayes (NB), Support Vector Classifier (SVC), k-nearest Neighbor (kNN), and Decision Tree (DT). Afterward, we perform classification with and without inferences using LSTM and BERT. We evaluate the models using the F1-score (see Subsection \ref{subsec:ExoSetyo}). We analyze the classification results by comparing the ten categories \ref{subsec:comAnaTenCates} and three types of classification methods \ref{subsec:comAnaThreAlgos}. 

\begin{table}
\centering
\caption{\bf F1-score of five different machine learning algorithms (LR, LSTM, BERT, PM-LSTM, and PM-BERT) for ten different categories (business, computers, games, health, home, recreation, science, shopping, society, and sports).}
\scalebox{0.68}{
\label{tab:resultsTable}
\begin{tabular}{|l|l|l|l|l|l|l|l|l|l|l|l|l|} 
\hline
\textbf{Model}                    & \textbf{Category} & \textbf{Cul} & \textbf{Eco} & \textbf{Pol} & \textbf{Ling} & \textbf{Geo} & \textbf{Category} & \textbf{Cul} & \textbf{Eco} & \textbf{Pol} & \textbf{Ling} & \textbf{Geo}  \\ 
\hline
\multirow{5}{*}{\textbf{LR}}      & Business          & 0.40              & 0.48              & 0.71               & 0.73                & 0.61                  & Recreation        & 0.37              & 0.30              & 0.59               & 0.73                & 0.57                   \\ 
\cline{2-13}
                                  & Computers         & 0.42              & 0.35              & 0.58               & 0.63                & 0.56                  & Science           & 0.42              & 0.42              & 0.62               & 0.71                & 0.65                   \\ 
\cline{2-13}
                                  & Games             & 0.52              & 0.35              & 0.59               & 0.59                & 0.60                  & Shopping          & 0.36              & 0.27              & 0.49               & 0.61                & 0.52                   \\ 
\cline{2-13}
                                  & Health            & 0.36              & 0.40              & 0.58               & 0.67                & 0.64                  & Society           & 0.40              & 0.45              & 0.62               & 0.68                & 0.62                   \\ 
\cline{2-13}
                                  & Home              & 0.39              & 0.57              & 0.49               & 0.68                & 0.59                  & Sports            & 0.44              & 0.28              & 0.59               & 0.62                & 0.57                   \\ 
\hline
\multicolumn{13}{|l|}{}                                                                                                                                                                                                                                                                          \\ 
\hline
\multirow{5}{*}{\textbf{LSTM}}    & Business          & 0.19              & 0.28              & 0.49               & 0.55                & 0.47                  & Recreation        & 0.20              & 0.39              & 0.48               & 0.41                & 0.47                   \\ 
\cline{2-13}
                                  & Computers         & 0.20              & 0.08              & 0.49               & 0.52                & 0.46                  & Science           & 0.20              & 0.20              & 0.48               & 0.43                & 0.43                   \\ 
\cline{2-13}
                                  & Games             & 0.14              & 0.29              & 0.74               & 0.70                & 0.48                  & Shopping          & 0.23              & 0.44              & 0.49               & 0.44                & 0.49                   \\ 
\cline{2-13}
                                  & Health            & 0.18              & 0.17              & 0.49               & 0.59                & 0.53                  & Society           & 0.20              & 0.27              & 0.49               & 0.46                & 0.48                   \\ 
\cline{2-13}
                                  & Home              & 0.19              & 0.21              & 0.49               & 0.59                & 0.63                  & Sports            & 0.26              & 0.43              & 0.48               & 0.43                & 0.49                   \\ 
\hline
\multicolumn{13}{|l|}{}                                                                                                                                                                                                                                                                          \\ 
\hline
\multirow{5}{*}{\textbf{BERT}}    & Business          & 0.42              & 0.54              & 0.62               & 0.74                & 0.50                  & Recreation        & 0.32              & 0.29              & 0.74               & 0.81                & 0.59                   \\ 
\cline{2-13}
                                  & Computers         & 0.38              & 0.30              & 0.47               & 0.75                & 0.66                  & Science           & 0.39              & 0.47              & 0.68               & 0.78                & 0.60                   \\ 
\cline{2-13}
                                  & Games             & 0.40              & 0.65              & 0.77               & 0.85                & 0.68                  & Shopping          & 0.34              & 0.44              & 0.49               & 0.63                & 0.50                   \\ 
\cline{2-13}
                                  & Health            & 0.38              & 0.54              & 0.66               & 0.79                & 0.67                  & Society           & 0.39              & 0.51              & 0.49               & 0.73                & 0.48                   \\ 
\cline{2-13}
                                  & Home              & 0.32              & 0.60              & 0.54               & 0.81                & 0.67                  & Sports            & 0.48              & 0.50              & 0.48               & 0.76                & 0.66                   \\ 
\hline
\multicolumn{13}{|l|}{}                                                                                                                                                                                                                                                                          \\ 
\hline
\multirow{5}{*}{\textbf{PM-LSTM}} & Business          & 0.19              & 0.28              & 0.49               & 0.43                & 0.48                  & Recreation        & 0.21              & 0.47              & 0.48               & 0.44                & 0.46                   \\ 
\cline{2-13}
                                  & Computers         & 0.21              & 0.06              & 0.49               & 0.46                & 0.47                  & Science           & 0.19              & 0.21              & 0.48               & 0.74                & 0.43                   \\ 
\cline{2-13}
                                  & Games             & 0.18              & 0.22              & 0.48               & 0.40                & 0.48                  & Shopping          & 0.18              & 0.50              & 0.49               & 0.50                & 0.48                   \\ 
\cline{2-13}
                                  & Health            & 0.20              & 0.27              & 0.48               & 0.43                & 0.46                  & Society           & 0.19              & 0.25              & 0.49               & 0.64                & 0.48                   \\ 
\cline{2-13}
                                  & Home              & 0.20              & 0.27              & 0.49               & 0.45                & 0.48                  & Sports            & 0.27              & 0.22              & 0.48               & 0.44                & 0.48                   \\ 
\hline
\multicolumn{13}{|l|}{}                                                                                                                                                                                                                                                                          \\ 
\hline
\multirow{5}{*}{\textbf{PM-BERT}} & Business          & 0.48              & 0.66              & \textbf{0.97}      & 0.80                & 0.96                  & Recreation        & \textbf{0.66}     & 0.35              & 0.74               & 0.48                & 0.65                   \\ 
\cline{2-13}
                                  & Computers         & 0.46              & 0.28              & 0.54               & \textbf{0.97}       & 0.70                  & Science           & 0.47              & 0.53              & \textbf{0.97}      & 0.78                & 0.66                   \\ 
\cline{2-13}
                                  & Games             & 0.33              & \textbf{0.72}     & 0.48               & 0.47                & 0.72                  & Shopping          & 0.45              & 0.50              & \textbf{0.97}      & 0.67                & 0.56                   \\ 
\cline{2-13}
                                  & Health            & 0.46              & 0.60              & 0.72               & 0.82                & \textbf{0.97}         & Society           & 0.52              & 0.69              & 0.55               & 0.73                & \textbf{0.97}          \\ 
\cline{2-13}
                                  & Home              & 0.41              & 0.66              & 0.54               & 0.83                & 0.73                  & Sports            & 0.49              & 0.50              & 0.54               & 0.79                & 0.63                   \\ 
\hline
\multicolumn{13}{|l|}{}                                                                                                                                                                                                                                                                          \\
\hline
\end{tabular}}
\end{table}

The results suggest that our proposed methods (LSTM and BERT with inferences based semantic knowledge and sentiments) outperform the three cultural, economic, political and geographical barrier.

\section{Conclusions}\label{sec:conclusions}
In this paper, we focused on the classification of news spreading barriers by utilizing the semantic knowledge in form of commonsense knowledge and sentiments. We consider the news related to the ten different categories (business, computers, games, health, home, recreation, science, shopping, society, and sports). After completing the automatic annotation of news data for the five barriers including cultural, economic, political, linguistic, and geographical (binary class classification of the linguistic, political, and geographical barrier and ternary class classification of the cultural and political barrier), we perform classification with classical machine learning methods (LR, NB, SVC, kNN, and DT), deep learning (LSTM) and transformer-based method (BERT). Our findings suggest that the commonsense based semantic knowledge and sentiments helps in achieving a higher F1-score.

\subsection{Theoretical and practical Implications}
The main theoretical contributions of this work are an approach to information barrier annotation based on news meta-data, and labelling and classifying the news including semantic knowledge across different barriers (cultural, economic, political, linguistic, and geographical).  
The annotation process includes meta-data extraction that requires too many requests to find the corresponding Wikipedia URL for a news publishers. Although many news publishers including some local news publishers do not have an entry to Wikipedia database, popular and a significantly huge amount of news publishers do have their profile available at Wikipedia-infobox. The annotation process and data statistics demonstrate that this approach of extracting profiles of news publishers is feasible to perform barrier classification to news spreading as well as for other important tasks such as understanding the fake news propagation. The used labeling process involve the demographic values and profile of news publishers such as cultural and economic differences, and political alignment and publishing language of news publishers. To the best of our knowledge, our proposed approach is first of its kind to classification of news spreading barriers.
\newline
In terms of practical contributions, there are basically two contributions: 1) an annotated data set, and 2) an approach to classification of news spreading barriers based on semantic knowledge including a wide range of common sense knowledge and sentiments of news headlines. Since the existing work lacks annotated dataset for this task, it presents an annotated data set for classification of news spreading barriers. It presents the sentiment analysis of annotated news headlines as well properties of the common sense knowledge relations in news headlines. Our experimental evaluation shows that the deep learning (LSTM) and transformer-based methods (BERT) can be useful for classifying the barriers using common sense based knowledge and sentiments.

\section{Acknowledgments}\label{sec:ack}
The research described in this paper was supported by the Slovenian research agency under the project J2-1736 Causalify
and by the European Union’s Horizon 2020 research and innovation program under the Marie Skłodowska-Curie grant agreement No 812997.

\bibliography{Main.bib}

\end{document}